\pdfoutput=1
\documentclass[7.8pt]{article}
\usepackage{acl}
\usepackage{times}
\usepackage{latexsym}
\usepackage[T1]{fontenc}
\usepackage[utf8]{inputenc}
\usepackage{microtype}
\usepackage{inconsolata}

\usepackage{tikz}
\usepackage{arydshln}
\usepackage{algpseudocode}
\algrenewcommand\textproc{\text}
\usepackage{makecell}
\usepackage{booktabs}
\usepackage{color}
\usepackage{colortbl}
\usepackage{multirow}
\usepackage{enumitem}
\usepackage{makecell}
\usepackage{threeparttable}
\usepackage{amsmath,amsfonts,mathtools}
\usepackage{pifont}

\usepackage{mathrsfs}
\usepackage{float}
\usepackage{graphicx}
\usepackage{xcolor}
\usepackage{enumitem}
\usepackage{tabularx, booktabs}

\usepackage{subfig}
\usepackage{tcolorbox}
\tcbuselibrary{breakable}
\definecolor{mygreen}{HTML}{00B050}

\definecolor{myorange}{HTML}{ED7D31}

\usepackage{array}
\usepackage{mathtools}
\usepackage{multirow}
\usepackage{subcaption}
\usepackage{tikz}
\renewcommand{\arraystretch}{1.1}
\definecolor{color1}{cmyk}{0.216,0.176,0,0}
\definecolor{color2}{cmyk}{0.059,0.235,0.392,0}

\usepackage{graphicx}
\usepackage{tikz}
\usepackage[ruled,vlined]{algorithm2e}
\usepackage{wrapfig}
\usepackage{algpseudocode}
\algrenewcommand\textproc{\text}
\usepackage{makecell}
\usepackage{booktabs}
\usepackage{pifont}
\usepackage{multirow}
\usepackage{enumitem}
\usepackage{balance}
\usepackage{threeparttable}
\usepackage{amsmath,amsfonts,mathtools} 
\usepackage{colortbl}
\usepackage{mathrsfs}
\usepackage{float}
\usepackage{graphicx}
\usepackage{hyperref}
\usepackage{tabularx, booktabs}
\usepackage{tikz}
\usepackage{arydshln}
\usepackage{CJKutf8}
\usepackage{graphicx}  
\usepackage{subfig}    

\definecolor{uc_color}{rgb}{0.99,0.24,0.63}
\definecolor{hc_color}{rgb}{0.02,0.51,0.51}
\definecolor{tc_color}{rgb}{0.99,0.55,0.09}

\tikzstyle{mybox} = [draw=black, very thick,
    rectangle, rounded corners, inner sep=10pt, inner ysep=13pt]
\tikzstyle{fancytitle} =[fill=black, text=white]

\newcommand{\M}{$3$\textsc{D}\textsc{S}}
\title{\M: Medical Domain Adaptation of LLMs via\\ \underline{D}ecomposed \underline{D}ifficulty-based \underline{D}ata \underline{S}election}

\author{
  Hongxin Ding\textsuperscript{1,2}\thanks{These authors contribute equally.}, 
  Yue Fang\textsuperscript{1,2}\footnotemark[1], 
  Runchuan Zhu\textsuperscript{1,2}\footnotemark[1], 
  Xinke Jiang\textsuperscript{1,2}, 
  Jinyang Zhang\textsuperscript{1,2}, \\
  \textbf{Yongxin Xu}\textsuperscript{1,2}, 
  \textbf{Weibin Liao}\textsuperscript{1,2}, 
  \textbf{Xu Chu}\textsuperscript{1,2}, 
  \textbf{Junfeng Zhao}\textsuperscript{1,2,4}\thanks{Corresponding authors.}, 
  \textbf{Yasha Wang}\textsuperscript{2,3,5}\footnotemark[2] \\
  \textsuperscript{1}School of Computer Science, Peking University, Beijing, China \\
  \textsuperscript{2}Key Laboratory of High Confidence Software Technologies, Ministry of Education; \\
  \textsuperscript{3}National Engineering Research Center for Software Engineering, Peking University, China \\
  \textsuperscript{4}Big Data Technology Research Center, Nanhu Laboratory, Jiaxing, China \\
  \textsuperscript{5}Peking University Information Technology Institute, Tianjin Binhai, China \\
  \texttt{\{dinghx, zhaojf, wangyasha\}@pku.edu.cn}
}

\begin{document}
\maketitle

\begin{abstract}
Large Language Models (LLMs) excel in general language tasks, motivating their adaptation to specialized domains such as healthcare. Effective domain adaptation typically involves supervised fine-tuning (SFT) on carefully selected instruction-tuning data. Current data selection methods adopt a \textbf{data-centric} approach, relying on external annotations and heuristics to identify externally defined high-quality or challenging data. 
Our exploratory experiments highlight this approach \textit{fails to improve the model's domain performance, due to misalignment between selected data and the model’s knowledge distribution}. To tackle this, we propose \underline{D}ecomposed \underline{D}ifficulty-based \underline{D}ata \underline{S}election (\textbf{\M}), a two-stage \textbf{model-centric} data selection framework that aligns data selection with the model’s distribution. 
\M~employs \textit{Prompt-Driven Data Selection} to filter out noise based on the model's knowledge via explicit alignment in Stage\#1, then adopts \textit{Decomposed Difficulty-based Data Selection} to guide selection via three novel data difficulty metrics, including \textit{Instruction Understanding}, \textit{Response Confidence}, and \textit{Response Correctness} in Stage\#2, enhanced by an \textit{attention-based importance weighting mechanism} for accurate calibration.
Extensive experiments in the healthcare domain show \M~outperforms existing methods by up to 2.97\% accuracy, with additional validation in law and general domains, confirming its generalization ability. Our dataset and code are open-sourced at \url{https://github.com/PuppyKnightUniversity/3DS}.
\end{abstract}


\section{Introduction}

Large Language Models (LLMs) such as proprietary GPT-4~\citep{OpenAI2023GPT4TR}, open-sourced LLaMA~\citep{touvron2023llama} and Qwen~\citep{bai2023qwen}, have demonstrated remarkable capabilities in language understanding and generation. Encouraged by their successes, there is growing interest in leveraging LLMs in specialized domains like healthcare, where domain-specific abilities are required~\citep{sanaei2023chatgpt,harris2023large,waisberg2023gpt} for essential tasks like diagnosis~\citep{panagoulias2024evaluating,ullah2024challenges,ding2025promed,fang2025eagrl}, treatment recommendations~\citep{wilhelm2023large,nwachukwu2024currently,timeline-yao2024overview}. To address this, many existing  works~\citep{Wang_Liu_Xi_Qiang_Zhao_Qin_Liu_2023,Zhang_Chen_Jiang_Yu_Chen_Li_Chen_Wu_Zhang_Xiao_et,Yang_Zhao_Zhu_Zhou_Xu_Jia_Zan_2023,Zhu_Togo_Ogawa_Haseyama,liao2025magical} have tried to adapt LLMs to the medical domain by training on large-scale healthcare-specific datasets.



An essential step in adapting general LLMs to specialized domains is Supervised Fine-Tuning (SFT) on domain instruction-tuning datasets. However, large-scale, unfiltered domain datasets aggregated from multiple sources often include \textit{noise}. Directly utilizing such data can disrupt learning~\citep{wang2023far, wang2024resilience}, hinder the identification of knowledge gaps~\citep{havrilla2024understanding}, and increase the risk of overfitting~\citep{budach2022effects, wang2024survey}, yielding poor performance. Recent findings~\citep{zhou2024lima} suggest that a \emph{small but carefully selected high-quality} dataset can effectively enhance model's alignment with instructions and elicit its abilities in the desired direction, highlighting the necessity of rigorous data selection for domain adaptation fine-tuning. This presents a critical challenge in fine-tuning general LLMs to specialized domains:

 \textit{How to identify and select domain instruction-tuning data that is most suitable for the target LLM to optimally elicit its domain-specific abilities?} 

Previous data selection methods predominantly adopt a \textbf{data-centric} perspective, typically focusing on two dimensions: \textit{quality} and \textit{difficulty}. For quality, existing methods rely on powerful external models or manual rules to identify ``high-quality'' samples~\citep{liu2023deita, ji2023survey, song2024typing}. They treat quality as a model-agnostic, intrinsic data property, assuming the assessments are universally applicable. However, LLMs differ substantially in architectures and training corpora, which shape their distinct internal knowledge distributions. External ``high-quality'' data may still introduce redundancy or conflicting information that impede learning. For difficulty, methods typically prioritize the most challenging samples based on heuristic metrics~\citep{li2024quantity, li-etal-2024-superfiltering}. However, recent studies~\citep{gekhman2024does, ren2024learning} have revealed that fine-tuning LLMs on data beyond their pre-trained knowledge distribution, particularly unfamiliar content, can lead to severe hallucinations, which underscores the potential risk of selecting hardest samples. A common limitation of these methods is their lack of consideration for model-specific compatibility, both external ``high-quality'' data or most challenging data could be misaligned with the model’s distribution and lead to suboptimal results.

Motivated by this gap, we propose a new hypothesis: \textit{data selection should be \textbf{model-centric}, tailored to align with the model’s knowledge distribution}. 

To validate this hypothesis, we conduct a pilot study guided by two research questions: \textbf{RQ\#1.} Is model-centric quality selection more effective than external quality scoring? \textbf{RQ\#2.} Is model-centric difficulty selection more effective than prioritizing the hardest samples? The results demonstrate that model-centric data selection, which relies on the target model’s own assessment of data quality and selection of appropriately difficult data, consistently outperforms selection guided by external criteria. 



While these findings highlight the importance of model-centric data selection, its practical application still faces substantial challenges:

\noindent\textbf{\ding{182} Challenge\#1. How to identify high-quality data based on the model's knowledge distribution?}
Redundant knowledge that the model already possesses and conflicting information against the model's knowledge hinders learning~\cite{ren2024learning,gekhman2024does}. Selecting high-quality data based on the model’s knowledge distribution is thus necessary, but inherently challenging due to the complexity and opacity of LLMs

\noindent\textbf{\ding{183} Challenge\#2. How to properly balance the selected data difficulty with the model's learning capacity?}
Overly simplistic data wastes training resources and may cause overfitting, while excessively complex data can overwhelm the model, impeding effective learning~\cite{kang2024unfamiliar,linflame,liao2025TPO}. Accurately assessing difficulty based on the model’s distribution to guide selection is thus crucial. However, there isn't an effective metric to comprehensively measure the model’s knowledge state and its ability to handle complex data.

To tackle these challenges, we propose \underline{\textbf{D}}ecomposed \underline{\textbf{D}}ifficulty-based \underline{\textbf{D}}ata 
\underline{\textbf{S}}election (\textbf{\M}), a two-stage \textbf{model-centric} data selection framework which aligns data selection with the model’s distribution to optimize domain fine-tuning. For \textbf{Challenge\#1}, we propose \textit{Prompt-Driven Data Selection via Explicit Alignment}, leveraging the target model's own evaluations to explicitly select high-quality data, ensuring that the remaining data lies within the model's knowledge distribution. For \textbf{Challenge\#2}, inspired by the general human problem-solving process~\cite{polya2014solve,/content/publication/PISA-creative-problem-solving-en}—understanding the problem, building confidence, and producing a solution, we propose novel \textit{Decomposed Difficulty-based Data Selection via Implicit Alignment}, extending traditional perplexity (PPL) measures with three difficulty metrics: Instruction Understanding Difficulty, Response Confidence Difficulty, and Response Correctness Difficulty. Furthermore, an \textit{attention-based importance weighting mechanism} captures token-level importance and calibrates difficulty calculations. In summary, our contributions are:
\begin{itemize}[leftmargin=*,noitemsep,topsep=2pt]
\item We introduce \M, a two-stage model-centric data selection framework, aligning training data with the model’s knowledge distribution, optimizing domain adaptation fine-tuning. 
\item We propose a novel difficulty decomposition strategy, employing fine-grained metrics: Instruction Understanding, Response Confidence, and Response Correctness, for accurate data difficulty quantification tailored to domain-specific fine-tuning.
\item Comprehensive experiments on Chinese medical datasets demonstrate that \M~outperforms existing methods, significantly boosting LLMs' performance. Additional experiments on law domain also showcase \M's~ generalization ability.
\item We have open-sourced a carefully curated Chinese medical dataset, including medical dialogues and domain-specific instructions, to support further research in healthcare-oriented LLM.
\end{itemize}

\section{Importance of Model-Centric Selection}
In this section, we empirically investigate the importance of model-centric data selection by studying the following two research questions:
\begin{itemize}[leftmargin=*,noitemsep,topsep=2pt]
\item \textbf{RQ\#1}. \textit{Is model-centric quality selection more effective than external quality scoring?}
\item \textbf{RQ\#2}. \textit{Is model-centric difficulty selection more effective than prioritizing the objectively hardest samples?}
\end{itemize}

\subsection{Experimental Setup}

In both investigations, we utilized two models: DeepSeek-R1~\cite{guo2025deepseek}, an external model regarded as strong and capable, which is expected to provide reliable data evaluation, and LLaMA3-8B-Instruct~\cite{grattafiori2024llama}, the target model intended for domain fine-tuning. We utilized a large-scale Chinese medical instruction-tuning dataset and designed tailored prompts to assess data quality and difficulty (see Appendix~\ref{appendix:quality_scoring_prompt} and \ref{appendix:difficulty_scoring_prompt}).


\subsection{Model-Centric \textit{vs.} External Quality Selection}

To answer \textbf{RQ\#1}, we prompted both the external model and the target model to assess data quality based on their knowledge. From data scored above a predefined threshold by each model, we randomly selected 5K samples and fine-tuned LLaMA3-8B-Instruct on each subset. Performance evaluated on two Chinese medical multiple-choice question benchmarks~\cite{zeng2023mmcu,wang2023cmb} is shown in Table~\ref{table:quality_empirical_study}.

\noindent Surprisingly, fine-tuning on high-quality data selected by the strong DeepSeek-R1 led to performance degradation of LLaMA3-8B-Instruct, while data selected by LLaMA3-8B-Instruct itself significantly improved its performance. This discrepancy likely stems from a misalignment between the external quality assessment and the target model's inherent knowledge distribution. Based on this, we derive our first key observation:


\noindent\textbf{Observation I:} \textit{Model-centric quality selection yields better performance than external quality scoring.}

\subsection{Model-Centric \textit{vs.} External Difficulty Selection}

To answer \textbf{RQ\#2}, we evaluated the commonly held assumption that training on the most challenging data improves model abilities. Similar to the previous investigation, we prompted DeepSeek-R1 and LLaMA3-8B-Instruct to score data difficulty based on their knowledge. The dataset was partitioned into Easy, Medium, and Hard subsets, according to difficulty scores from each model. We then fine-tuned LLaMA3-8B-Instruct on randomly selected 5K samples from each subset, and compared their performance across medical benchmarks, with results shown in Table~\ref{table:difficulty_empirical_study}.

\noindent Fine-tuning on \textit{Easy} and \textit{Medium} subsets consistently outperformed training on \textit{Hard} subset, with Medium subset yielding more stable improvements. This indicates that overly difficult data likely exceeding model's knowledge, adversely impacts learning, while overly simple data fails to sufficiently benefit fine-tuning. Additionally, difficulty assessments from LLaMA3-8B itself consistently led to better results compared to external evaluations by DeepSeek-R1, which validates the necessity of model-centric difficulty evaluation and selection. This motivates our second and third key observations:

\noindent\textbf{Observation II:} \textit{Difficulty scoring based on the target model yields more reliable  performance than scores provided by an external model.}

\noindent\textbf{Observation III:} \textit{Moderately difficult data leads to more stable and effective performance improvements.}

\begin{table}[htbp]
\centering
\resizebox{\columnwidth}{!}{
{ \renewcommand{\arraystretch}{0.9}  
\begin{tabular}{lccc}
\toprule
\textbf{Data} & \textbf{Annotator} & \textbf{CMB-Exam} & \textbf{MMCU-Med} \\ 
\midrule
No SFT & N/A & 41.72 & 46.47 \\
\midrule
\multirow{2}{*}{High-quality}
& DeepSeek-R1 & 39.70 & 42.46 \\
& LLaMA3-8B & \textbf{43.71} & \textbf{47.57} \\
\bottomrule
\end{tabular}
}
}
\caption{High-quality Data Selection Results (\%). Improvements over the original model are in \textbf{bold}.}
\label{table:quality_empirical_study}
\end{table}

\vspace{-10pt}

\begin{table}[htbp]
\centering
\resizebox{\columnwidth}{!}{
{ \renewcommand{\arraystretch}{0.9}  
\begin{tabular}{lccc}
\toprule
\textbf{Data} & \textbf{Annotator} & \textbf{CMB-Exam} & \textbf{MMCU-Med} \\ 
\midrule
No SFT & N/A & 41.72 & 46.47 \\
\midrule
\multirow{2}{*}{Easy}
& DeepSeek-R1 & 41.03 & 45.76 \\
& LLaMA3-8B & 41.53 & \textbf{48.00} \\
\midrule
\multirow{2}{*}{Medium}
& DeepSeek-R1 & \textbf{41.76} & 45.26 \\
& LLaMA3-8B & \textbf{41.75} & \textbf{46.72} \\
\midrule
\multirow{2}{*}{Hard}
& DeepSeek-R1 & 40.50 & 44.06 \\
& LLaMA3-8B & 40.62 & 45.23 \\
\bottomrule
\end{tabular}
}
}
\caption{Difficult Data Selection Results (\%). Improvements over the original model are in \textbf{bold}.
}
\label{table:difficulty_empirical_study}
\end{table}
\vspace{-8pt}

\subsection{Conclusion and Motivation}
Both investigations lead to a key conclusion: effective data selection for domain adaptation fine-tuning requires alignment with the target model's knowledge distribution. External assessed high-quality data may not suit the target model, and excessively difficult data may introduce unfamiliar, out-of-distribution content, causing suboptimal outcomes.

Motivated by these observations, we propose to shift from conventional \textbf{data-centric} selection strategies toward a \textbf{model-centric} approach. Specifically, data selection should be guided by the target model, ensuring that the selected data are considered as high-quality (addressing \textbf{Observation I}) and appropriately challenging by the target model (addressing \textbf{Observation II and III}), thus achieving close alignment with its knowledge distribution and learning capacity. Building on this insight, we propose our novel model-centric framework \M~in the following sections.

\section{Methodology}
\label{method}

\paragraph{Task Formulation}
We formally define the Data Selection for Domain Adaptation Fine-tuning task. Let:
\begin{itemize}[leftmargin=*,noitemsep,topsep=2pt]
    \item $M_{\theta}$ denotes the target model to be fine-tuned, which is a pre-trained and generally fine-tuned LLM (e.g., LLaMA-chat) parameterized by $\theta$.
    \item $\mathcal{X}=\{x^{(i)}\}^N_{i=1}$ denotes the full domain-specific dataset where each sample $x^{(i)}=<Q^{(i)},A^{(i)}>$ consists of instruction $Q^{(i)}=\{q^{(i)}_1,q^{(i)}_2,\dots,q^{(i)}_m\}$, and response $A^{(i)}=\{a^{(i)}_1,a^{(i)}_2,\dots,a^{(i)}_n\}$. Here $q^{(i)}_m, a^{(i)}_n$ denote individual tokens within the instruction and response sets, respectively.
    \item $k\in \mathbb{N}^+$ denotes a fixed data budget, where $k \ll |\mathcal{X}|$.
\end{itemize}
The task is to identify an optimal subset $\mathcal{S^*}\subseteq \mathcal{X}$ that maximizes the target domain performance of the fine-tuned model $M_{\theta}'$, formally:
\begin{equation}
\begin{aligned}
    \mathcal{S^*}=\mathop{\arg\max}\limits_{S \subseteq \mathcal{X}, |S| = k} \mathbb{E}_{(x,y) \sim \mathcal{D}_{\text{test}}} \left[ \mathcal{P}(M_{\theta'}(x; \mathcal{S}), y) \right],
\end{aligned}
\end{equation}
where $\mathcal{D}_{test}$ is the target domain test distribution containing diverse tasks; $\mathcal{P}: \mathcal{Y}\times\mathcal{Y}\xrightarrow{}[0,1]$ is the performance metric (e.g., accuracy, BLEU), and $M_{\theta'}$ is $M_{\theta}$ fine-tuned on $\mathcal{S}$, i.e., $\theta'=\theta - \eta \nabla_{\theta} \sum_{x \in S} \mathcal{L}(M_{\theta}(x), x)$, with learning rate $\eta$ and loss function $\mathcal{L}$.
\begin{figure*}[ht]
    \centering
    \includegraphics[width=1\linewidth]{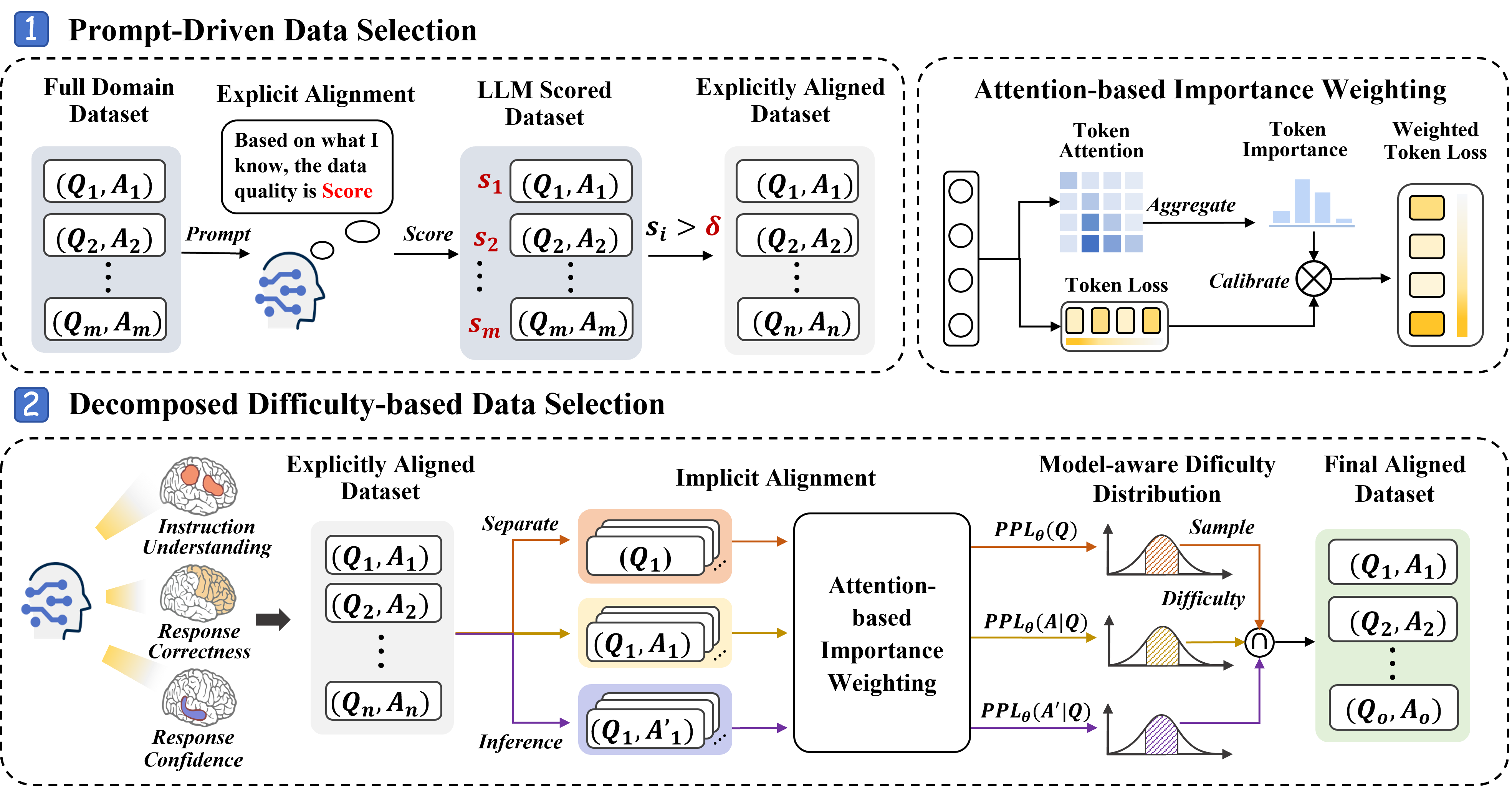}
    \caption{\textbf{\M~framework}. 
    \textbf{Stage\#1:} Prompt-Driven Data Selection selects high-quality data via explicit alignment. 
    \textbf{Stage\#2:} Decomposed Difficulty-based Data Selection decomposes data difficulty via modeling LLM's implicit distribution and filters data. 
    Attention-based importance weighting calibrates difficulty calculation.}
    \label{fig:framework_overview}
\end{figure*}

\subsection{Stage\#1: Prompt-Driven Data Selection via Explicit Alignment}
The first stage of \M~is to identify high-quality data based on the model's knowledge. As illustrated in Figure~\ref{fig:framework_overview}, a quality-rating prompt, detailed in Appendix~\ref{appendix:quality_scoring_prompt}, is used to instruct $M_{\theta}$ to score data quality based on its inner knowledge to explicitly align data, filtering out noise from the original large-scale dataset to avoid conflicting information. After obtaining model-generated scores, samples with scores exceeding a predefined threshold $\delta$ are retained for the next selection. 





\subsection{Stage\#2: Decomposed Difficulty-based Data Selection via Implicit Alignment}
The second stage of \M~is to analyze data difficulty via implicit distribution modeling of $M_{\theta}$, thereby balancing the selected data difficulty with the model's learning capacity. To achieve this, we employ a fine-grained evaluation for data difficulty. 

Inspired by the general problem-solving process~\cite{polya2014solve,/content/publication/PISA-creative-problem-solving-en}—understanding the problem, building confidence, and producing a solution—we decompose data difficulty into three key components to reflect the model's understanding: (1) \textit{Instruction Understanding Difficulty} measures whether the model comprehends the instruction. (2) \textit{Response Confidence Difficulty} measures the model's confidence in its response. (3) \textit{Response Correctness Difficulty} measures whether the model can generate a response that accurately matches the reference answer. To enhance the precision of difficulty calculations, we incorporate an \textit{attention-based importance weighting mechanism} that calibrates difficulty by accounting for the varying semantic significance of output tokens. We now detail the quantification of these decomposed difficulties and the corresponding selection strategy. 

\paragraph{(1) Instruction Understanding Difficulty.} Challenging data often comes with complex instructions. In specialized domains like healthcare, instructions may contain intricate terminologies, making instruction comprehension a key factor of data difficulty. To capture this, we introduce Instruction Understanding Difficulty. Previous research \citep{gonen-etal-2023-demystifying} shows that lower model perplexity over a prompt correlates with better understanding and performance. Building on this insight, we further recognize that perplexity inherently captures the predictive uncertainty from model's distribution. Consequently, we employ model perplexity as a measure to quantify data difficulty from the model’s perspective. Formally, for a model $M_{\theta}$, given a data sample $x=<Q,A>$ with instruction $Q=\{q_1,q_2,\dots q_m\}$, its Instruction Understanding Difficulty is defined as:
\begin{equation}\label{D1}
\begin{aligned}
&\text{D1}_{\theta}(x)=\text{PPL}_{\theta}(Q)\\ &= \exp\left( -\frac{1}{m} \sum_{i=1}^{m} \log P_{\theta}(q_i | q_1, q_2, \dots, q_{i-1}) \right),
\end{aligned}
\end{equation}
where $P_{\theta}(q_i | q_1, q_2, \dots, q_{i-1})$ represents the probability $M_{\theta}$ generates the $i$-th token in instruction $Q$ given the preceding tokens. Higher perplexity indicates greater difficulty for the model to comprehend the instruction.

\paragraph{(2) Response Confidence Difficulty.} When encountering challenging data, models often struggle to provide a confident response. This uncertainty arises from its inability to handle the task and determine the most appropriate response, similar to human learners~\cite{preheim2023assessing_confidences}, which indicates high data difficulty. To quantify this difficulty, we introduce Response Confidence Difficulty, measured by the model's conditional perplexity when generating a response given the instruction. Formally, for a model $M_{\theta}$, given a data sample $x=<Q,A>$ with instruction $Q$ is and model-generated response $A'=\{a_1',a_2'\dots,a_{n'}'\}$ based on $Q$, its Response Confidence Difficulty is defined as:
\begin{equation}\label{D2}
 \begin{aligned}
    &\text{D2}_{\theta}(x) = \text{PPL}_{\theta}(A'|Q)\\ &=\exp\left( -\frac{1}{n'} \sum_{j=1}^{n'} \log P_{\theta}(a_j' | a_1', a_2', \dots, a_{j-1}',Q) \right).
\end{aligned}
\end{equation}
Higher conditional perplexity indicates greater uncertainty in the model's distribution and greater difficulty for the model to provide a confident answer.

\paragraph{(3) Response Correctness Difficulty.} For instruction-tuning data with reference answers, it is essential to assess the model's ability to generate correct responses to assess data difficulty. We introduce Response Correctness Difficulty, measured by the model's conditional perplexity when generating the reference answer $A=\{a_1,a_2\dots,a_n\}$ given instruction $Q$. 
\begin{equation}\label{D3}
\begin{aligned}
    &\text{D3}_{\theta}(x) = \text{PPL}_{\theta}(A|Q)\\ &= \exp\left( -\frac{1}{n} \sum_{j=1}^{n} \log P_{\theta}(a_j | a_1, a_2, \dots, a_{j-1}, Q) \right).
\end{aligned}
\end{equation}
Higher conditional perplexity indicates greater difficulty in producing the correct response, suggesting the sample poses more challenge for the model.

\paragraph{Attention-based importance weighting mechanism.} Response Confidence and Response Correctness Difficulties rely on evaluating the uncertainty inherent in the model's generation process. While conditional perplexity serves as an effective proxy, it treats all tokens equally, disregarding their varying semantic importance. While key tokens significantly influence the meaning and correctness of a response, trivial tokens like conjunctions may exhibit high uncertainty without substantially influencing semantics. This can lead to inaccurate data difficulty assessments. To address this, inspired by \citet{su2024dragin}, we introduce an attention-based importance weighting mechanism that adjusts token's uncertainty contributions by weighting based on their semantic importance. We argue that critical tokens are those playing a pivotal role in guiding subsequent generations. Therefore, we derive importance scores from the model's internal attention mechanism. Specifically, for a token sequence $s=\{t_1,t_2,\dots,t_i,\dots,t_n\}$, when a transformer-based LLM generates token $t_j(i<j)$, it computes the attention weight $A_{ji}$ by applying a softmax function to the dot product of the query vector $q_{j}$ and the key vector $k_{i}$:\begin{equation}
\begin{aligned}
     A_{ji} = \texttt({q_j \cdot k_i})/{\sqrt{d_k}},
\end{aligned}
\end{equation}
where $d_k$ is the dimension of $k_{i}$. $A_{ji}$ represents the attention the model pays to token $t_i$ when generating token $t_j$, reflecting the importance of $t_i$. We define the importance score of token $t_i$ as the aggregated attention weight it receives from all subsequent tokens:
\begin{equation}
\begin{aligned}
    \text{I}(t_i) = \underset{j > i}{\texttt{Aggregate~}} (A_{ji}).
\end{aligned}
\end{equation}
We use mean aggregation to compute token importance scores. Using these scores, Response Confidence and Response Correctness Difficulties are refined as:

\begin{equation}
\begin{aligned}
    \texttt{Atten-D2}_{\theta}(x) &= \texttt{weightedPPL}_{\theta}(A'|Q) \\&=
    \exp\left( - \frac{\sum_{j=1}^{n'} \text{I}(t_j) \cdot \phi_j'}{\sum_{j=1}^{n'} \text{I}(t_j)} \right), \\
    \phi_j' &= \log P_{\theta}(a_j' | a_1', a_2', \dots, a_{j-1}', Q),
\end{aligned}
\end{equation}


\begin{equation}
\begin{aligned}
    \texttt{Atten-D3}_{\theta}(x) &= \texttt{weightedPPL}_{\theta}(A|Q) \\ &= 
     \exp\left( - \frac{\sum_{j=1}^{n} \text{I}(t_j) \cdot \phi_j}{\sum_{j=1}^{n} \text{I}(t_j)} \right), \\
     \phi_j &= \log P_{\theta}(a_j | a_1, a_2, \dots, a_{j-1}, Q).
\end{aligned}
\end{equation}

By integrating attention-based importance weights, this mechanism prioritizes tokens crucial for semantic correctness and clarity, offering a more accurate estimation of model uncertainty and data difficulty. 

\paragraph{Selection Strategy based on Decomposed Difficulty.} Based on the decomposed data difficulties, \M~identifies samples whose difficulty metrics fall within a predefined middle range, discarding either trivially easy or overly complex data, focusing on moderately challenging samples that match the model's learning capacity. K-Center sampling (introduced in Appendix~\ref{appendix:kcenter}) based on instruction embeddings is then applied on this subset to enhance data diversity, reducing the risk of overfitting on highly similar samples.

\subsection{Model-Centric Data Selection Framework}
The overall architecture of our model-centric data selection framework is illustrated in Figure~\ref{fig:framework_overview}. Pseudo codes of the process are shown in Appendix~\ref{appendix:pseudo_codes}.

\section{Main Experiments}

\begin{table*}[h]
\label{main_result1}
\setlength{\tabcolsep}{1mm}
\centering
\resizebox{\linewidth}{!}
{
\begin{tabular}{cc|cc|cc|cc|cc|c}
\hline
\multirow{2}{*}{\textbf{Method}} & \multicolumn{1}{c|}{\textbf{LLM Turbo}} & \multicolumn{2}{c|}{\textbf{Baichuan2-13B-Chat}}& \multicolumn{2}{c|}{\textbf{Qwen1.5-7B-Instruct}} & \multicolumn{2}{c|}{\textbf{Qwen2.5-7B-Instruct}} & \multicolumn{2}{c|}{\textbf{LLaMA3-8B-Instruct}} & \multirow{2}{*}{\textbf{Avg.}} \\
\cline{2-10} 
& \textbf{Dataset} & CMB-Exam & MMCU-Med  & CMB-Exam & MMCU-Med & CMB-Exam & MMCU-Med  & CMB-Exam & MMCU-Med \\

\hline
\multirow{8}{*}{\textbf{Baselines}} 
& Base    & 46.67 & 47.11 & 59.80 & 64.24 & 78.28 & 83.43 & 41.72 & 46.47 & 58.47\\
& Full-SFT & 40.38 & 37.90 & 48.05 & 47.53 & 71.04 & 75.49 & 40.85 & 46.72 & 51.00\\
& Random   & 44.07 & 47.61 & 61.81 & 65.10 & 75.92 & 82.41 & 41.54 & 45.23 & 57.96\\

& Alpagasus & 42.24 & 43.56 & 55.67 & 58.74 & 69.90 & 78.08 & 41.60 & 45.26 & 54.38\\
& DEITA     & 46.78 & 49.88 & 45.33 & 44.09 & 74.07 & 81.59 & 41.31 & 45.80 & 53.60\\
& MoDS      & 47.25 & 50.37 & 61.09 & 64.67 & 76.31 & 82.23 & 39.25 & 42.53 & 57.96\\
& IFD       & 46.44 & 50.08 & \textbf{62.06} & 65.37 & 78.17 & 84.57 & 38.25 & 40.48 & 58.18\\
& LESS      & 45.79 & 51.01 & 60.74 & 64.85 & 78.83 & 83.20 & 41.80 & 44.63 & 58.86\\

\hline
\textbf{Ours}
 & \textbf{\M} & \textbf{47.37} & \textbf{51.08} & 61.96 & \textbf{66.09} & \textbf{79.06} & \textbf{85.70} & \textbf{43.95} & \textbf{49.70} & \textbf{60.61}\\
\hline
\rowcolor[gray]{0.95} \multicolumn{2}{c|}{*\textbf{Performance Gain $\uparrow$}}   &0.70  &3.97 &2.16 &1.85 & 0.78 & 2.27 & 2.23 & 3.23 & 2.14\\
\hline
\end{tabular}
}
\caption{Performance comparison (\%) on CMB-Exam and MMCU-Med of EM score. The best performance is highlighted in \textbf{bold}. Performance gains are measured against the base model.
}
\label{tab:comparisonQA}
\end{table*}


\begin{table*}[h]
\label{main_result2}
\setlength{\tabcolsep}{1mm}
\centering
\resizebox{\linewidth}{!}
{
\begin{tabular}{cc|ccc|ccc|ccc|ccc|c}
\hline
\multirow{2}{*}{\textbf{Method}} & \multicolumn{1}{c|}{\textbf{LLM Turbo}} & \multicolumn{3}{c|}{\textbf{Baichuan2-13B-Chat}}& \multicolumn{3}{c|}{\textbf{Qwen1.5-7B-Instruct}} & \multicolumn{3}{c|}{\textbf{Qwen2.5-7B-Instruct}} & \multicolumn{3}{c|}{\textbf{LLaMA3-8B-Instruct}}& \multirow{2}{*}{\textbf{Avg.}}\\
\cline{2-14} 
& \textbf{Metric} & BLEU-1 & BLEU-4  & ROUGE & BLEU-1 & BLEU-4  & ROUGE & BLEU-1 & BLEU-4  & ROUGE & BLEU-1 & BLEU-4  & ROUGE \\

\hline
\multirow{8}{*}{\textbf{Baselines}} 
& Base  &11.15  &21.02 &14.08 &16.17 &32.03 &16.31 & 21.87 & 64.11 & 36.74 & 5.06 & 35.09 & 10.40 & 23.67\\
& Full-SFT &7.19  &16.33 &11.70 &6.68 &16.61 &9.62 & 16.72 & 36.52 & 19.84 & 2.80 & 6.87 & 6.58 & 13.12\\
& Random  &12.14  &25.95 &14.75 &16.09 &34.45 & 16.19 & 16.49 & 33.68 & 17.89 & 9.01 & 25.49 & 12.14 & 19.52\\

& Alpagasus  & 10.16 & 20.42 & 12.58 & 14.48 & 31.63 & 14.77 & 16.85 & 35.74 & 18.77 & 8.66 & 22.51 & 12.36 & 18.24\\
& DEITA &19.42 & 42.07 & 19.32 & 18.92 & 42.93 & 20.32 & 21.71 & 49.33 & 23.40 & 9.91 & 23.33 & 13.86 & 25.38\\
& MoDS  &22.43  &51.02 &22.85 &17.61 &39.19 &19.93 & 18.83 & 41.31 & 21.45 & 12.38 & 29.74 & 15.33 & 26.01\\
& IFD &21.44  &51.73  &24.94 &19.24 &43.10 &21.08 & 18.07 & 39.16 & 20.28 & 10.59 & 29.32 & 14.83 & 26.15\\
& LESS  &13.27  &29.20 &16.40 &17.48 &38.88 &17.58 & 19.08 & 45.20 & 22.42 & 11.82 & 31.98 & 15.55 & 23.24\\

\hline
\multirow{1}{*}{\textbf{Ours}} 
 & \textbf{\M} &\textbf{24.15}  &\textbf{63.51} &\textbf{31.50} & \textbf{24.40} & \textbf{60.32} &\textbf{28.07} & \textbf{22.05} & \textbf{64.95} & \textbf{37.11} & \textbf{12.52} & \textbf{36.88} & \textbf{17.09} & \textbf{35.21} \\

\hline
\rowcolor[gray]{0.95} \multicolumn{2}{c|}{*\textbf{Performance Gain $\uparrow$}}  &13.00  &42.49  &17.42 &9.45 &29.49 &11.92 & 0.18 & 0.84 & 0.37 & 7.46 & 1.79 & 6.69 & 11.54\\ \hline
\end{tabular}
}
\caption{Performance comparison (\%) on CMB-Clin. 
The best performance is highlighted in \textbf{bold}. Performance gains are measured against the base model.
}
\label{tab:comparisonQAclin}
\end{table*}

\subsection{Experimental Setup}
\paragraph{Training dataset.} For medical domain adaptation fine-tuning, we construct a comprehensive medical instruction-tuning dataset of diversity and abundance. The dataset comprises over 1.9M samples, with its statistics provided in Table~\ref{tab:train_dataset_statistics} and data construction details introduced in Appendix~\ref{appendix:train_dataset}. We have released this complete training dataset to support further research.

\paragraph{Evaluation datasets.}
We assess models on diverse medical test datasets: two multi-task, multiple-choice datasets, MMCU-Med \citep{zeng2023mmcu} and CMB-Exam \citep{wang2023cmb}, and an open Q\&A dataset, CMB-Clin \citep{wang2023cmb}. Data statistics are provided in Table~\ref{tab:test_dataset_statistics}. MMCU-Med and CMB-Exam, consisting of medical exam questions, assess the model's reasoning and medical knowledge application abilities with accuracy as the metric. CMB-Clin, comprising patient record analysis tasks, assesses the model's complex medical analysis ability, with BLEU-1, BLEU-4 and ROUGE as the metric (detailed in Appendix~\ref{appendix:metrics}). Together, these datasets provide a comprehensive evaluation of the model's proficiency in the medical domain.

\paragraph{Models.} Experiments are conducted on instruct models of varying architectures and parameter sizes: Baichuan2-13B-Chat~\citep{baichuan}, Qwen1.5-7B-Instruct, Qwen2.5-7B-Instruct~\citep{bai2023qwen} and LLaMA3-8B-Instruct~\citep{touvron2023llama}.

\paragraph{Baselines.} We compare \M~with a series of LLM fine-tuning data selection strategies. 
(1)~\textbf{Base} directly tests models without further fine-tuning. 
(2)~\textbf{Full-SFT} fine-tunes models on the full training set. 
(3)~\textbf{Random Selection} randomly selects data. (4)~\textbf{Alpagasus}~\cite{chen2023alpagasus} utilizes GPT-4 to identify high-quality data. (5)~\textbf{DEITA}~\cite{liu2023deita} trains quality and complexity scorers and selects data according to their judgments (6)~\textbf{MoDS}~\citep{du2023mods} filters high-quality data via a reward model, and selects data necessary for model learning through training and inference processes. (7)~\textbf{IFD}~\citep{li-etal-2024-superfiltering,li2024quantity} designs instruction following difficulty metric based on the ground truth output loss with or without inputs. 
(8)~\textbf{LESS}~\citep{xia2024less} searches for training data similar to target task examples through low-rank gradient similarity. The implementation details are introduced in Appendix~\ref{appendix:baseline}.

\paragraph{Implementations.} The data budget is 5K samples. In \M, Prompt-Driven Data Selection retains samples with a quality score $\ge90$. In subsequent Decomposed Difficulty-based Data Selection, difficulty thresholds are determined via experiments on CMB hold-out validation set. Specifically, for Baichuan2-13B-Chat, the thresholds are set to 15\% and 65\%; for Qwen1.5-7B-Instruct and Qwen2.5-7B-Instruct, 25\% and 75\%; and 40\% and 90\% for LLaMA3-8B-Instruct. More 
implementation details are introduced in Appendix~\ref{appendix:hyperparameters}.

\subsection{Main Results}
Experiment results are shown in Table~\ref{tab:comparisonQA} and Table~\ref{tab:comparisonQAclin}. We summarize our findings below. 

\paragraph{Data selection is necessary for LLM domain adaptation fine-tuning.} We observe that fine-tuning LLMs on the full 1.9 million dataset (Full-SFT) leads to drastic performance drops across three benchmarks. This suggests that domain datasets directly collected from public resources contain significant noise that hinders model learning, highlighting the necessity of data selection. 

\paragraph{\M~effectively enhances LLM's diverse domain abilities, significantly outperforming baselines.} As shown in Table~\ref{tab:comparisonQA} and Table~\ref{tab:comparisonQAclin}, across various benchmarks and LLM backbones, \M~generally achieves the highest accuracy, outperforming the backbones and strong data selection baselines. On medical exam datasets, it improves base model performance by up to 8.43\% (on MMCU-Med for Baichuan2-13B-Chat), and exceeds the best baseline an average of 2.97\%, greatly enhancing the model's medical knowledge application abilities. On the open Q\&A CMB-Clin, models fine-tuned with \M~significantly outperforms all baselines by a large margin, exhibiting superior medical analysis ability.
To more comprehensively analyze model's domain performance, for CMB-Clin, we also conduct a pair-wise comparison using GPT-4o as the judge, detailed in Appendix~\ref{appendix:win_rates}. Both the quantitative and qualitative evidence demonstrate that \M~boosts the model's multi-faceted domain abilities. 

In contrast, methods relying on external, model-agnostic data evaluations, such as Alpagasus and DEITA, often lead to performance declines, especially on Qwen models. This further validates our previous conclusion that misalignment between selected data and the model hinders learning. Baselines MoDS and IFD show relatively strong results due to their considerations of model distribution and data difficulty. However, their selection on the most challenging data also proves inefficient, as they only bring marginal improvements across tasks and even underperform the backbone and random selection on LLaMA3-8B-Instruct. Baseline LESS, which aims to enhance performance on one specific downstream task, fails to generalize to domain adaptation fine-tuning where diverse abilities need to be improved, leading to performance degradation on MMCU-Med for Baichuan2-13B-Chat.

\paragraph{\M~exhibits strong generalization ability.} \M's consistent performance gains across backbones and benchmarks highlight its generalization ability to adapt to different models and domain tasks. To further validate the practicality of \M, we compare models fine-tuned using \M-with existing medical LLMs, with results shown in Appendix~\ref{appendix:comparison_with_medical_LLMs}.

\subsection{Ablation Studies}
\label{ablation_studies}
\M~is composed of two stages, and in Stage\#2, three difficulty metrics are proposed. To validate the effectiveness of each component, we conduct comprehensive ablation studies. Without loss of generality, experiments are done on Baichuan2-13B-Chat and Qwen1.5-7B-Instruct. Main results are shown in Table~\ref{tab:main_ablations}, and additional metrics are in Appendix~\ref{appendix:more_results_for_3DS_ablation}.

\begin{table*}[ht]
\setlength{\tabcolsep}{1mm}
\centering
\resizebox{0.95\linewidth}{!}{
\begin{tabular}{c|ccc|ccc|c}
\hline
\multicolumn{1}{c|}{\textbf{LLM Turbo}} & \multicolumn{3}{c|}{\textbf{Baichuan2-13B-Chat}}& \multicolumn{3}{c|}{\textbf{Qwen1.5-7B-Instruct}} & \multirow{2}{*}{\textbf{Avg.}} \\
\cline{1-7} 
\textbf{Benchmark} & CMB-Exam & MMCU-Med & CMB-Clin & CMB-Exam & MMCU-Med & CMB-Clin  \\
\hline
\textbf{w/o Stage\#1} & 44.64 & 48.06 & 16.19 & 60.37 & 64.03 & 15.88 & 41.53\\
\textbf{w/o Stage\#2} & 47.09 & 50.83 & 21.83 & 61.59 & 65.91 & 21.55 & 44.80\\
\textbf{Stage\#2 Collapsed into Stage\#1} & 47.28 & \underline{51.01} & 22.69 & 60.56 & 63.99 & 23.41 & 44.82\\
\hline
\textbf{w/o D1} & \underline{47.35} & 50.59 & 23.99 & 61.47 & 65.80 & 24.68 & 45.65\\
\textbf{w/o D2} & 47.34 & 47.18 & 23.54 & \textbf{62.00} & \underline{66.05} & 23.84 & 44.99\\
\textbf{w/o D3} & 47.07 & 50.59 & 23.08 & 61.64 & 65.73 & 23.83 & 45.32\\
\textbf{w/o Atten} & 47.10 & 50.19 & \underline{29.58} & 61.79 & 65.84 & \underline{27.69} & 47.03\\ 
\hline
\rowcolor[gray]{0.95}\textbf{\M} & \textbf{47.37} & \textbf{51.08} & \textbf{31.50} & \underline{61.96} & \textbf{66.09} & \textbf{28.07} & \textbf{47.68}\\
\hline
\end{tabular}
}
\caption{Performance comparisons (\%) on CMB-Exam, MMCU-Med and CMB-Clin of ablation studies on stages and difficulty metrics of \M. For CMB-Clin, the ROUGE score is reported. 
}
\label{tab:main_ablations}
\end{table*}

\paragraph{Ablation on stages.}
To evaluate the contributions of each stage in \M, we compare: \textbf{(1) removing Stage\#1}, where 70K samples are randomly sampled from the complete training dataset for subsequent difficulty-based selection, and \textbf{(2) removing Stage\#2}, where K-Center sampling is directly applied to the high-quality samples identified in stage\#1. Additionally, to validate the necessity of decomposed difficulty calculation based on model perplexity, we investigate \textbf{(3) collapsing Stage\#2 into Stage\#1}, where the model is prompted to verbalize its assessments of the three data difficulties (corresponding prompts are shown in Appendix~\ref{appendix:decomposed_difficulty_prompts}), bypassing the original difficulty calculation.

The results show a consistent pattern: \textbf{each modification leads to a decrease in performance compared to the original method}, emphasizing the necessity of quality controls and selecting appropriately difficult data. When Stage\#2 is collapsed into Stage\#1 via difficulty evaluation prompts, performance also degrades. During experiments, we observe that LLMs struggle to provide fine-grained assessments of data difficulty, often generating coarse-grained scores such as 0.5, 0.8, and 1. This lack of granularity makes it challenging to identify nuanced differences in data difficulty and select targeted data with desired moderate difficulties. 

While results on multiple-choice benchmarks do not indicate which stage is more important, the analysis performance on CMB-Clin reveals a clearer trend: removing Stage\#1 leads to the poorest performance, followed by removing Stage\#2 and collapsing Stage\#2 into Stage\#1. This pattern highlights the crucial role of quality control for the model to provide coherent and high-quality answers. Difficulty-based selection is also essential, as even coarser-grained difficulty measurements by model verbalization yield better results than ignoring difficulty at all. This progressive improvement further reinforces the two-stage design of \M.

\paragraph{Ablation on difficulty metrics.}
We remove each of the three metrics, Instruction Understanding, Response Confidence, and Response Correctness and run \M~without any other modifications. The results in Table~\ref{tab:main_ablations} demonstrate that, in general, removing any single component results in noticeable performance drops, indicating a decline in certain aspects of the model's medical abilities. These observations validate the necessity of each difficulty metric in identifying beneficial data samples for enhancing LLM's domain abilities. Additionally, removing the attention-based importance weighting mechanism also brings performance declines, which validates its effectiveness.

Additional ablation studies on data budgets are introduced in Appendix~\ref{appendix:ablation on data budgets}.

\subsection{Impact of Difficulty Thresholds}
\begin{figure}[ht]
    \centering
    \subfloat[Baichuan2-13B-Chat]{\includegraphics[width=0.48\columnwidth]{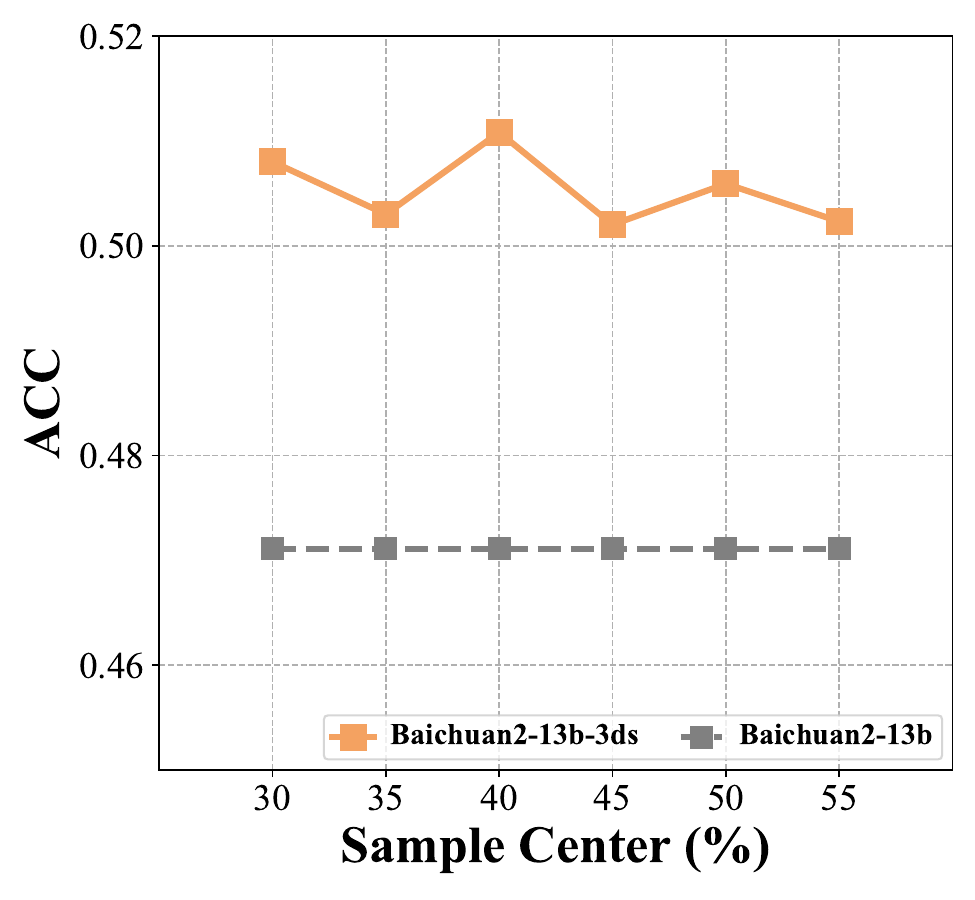}}
    \hfill
    \subfloat[Qwen1.5-7B-Instruct]{\includegraphics[width=0.48\columnwidth]{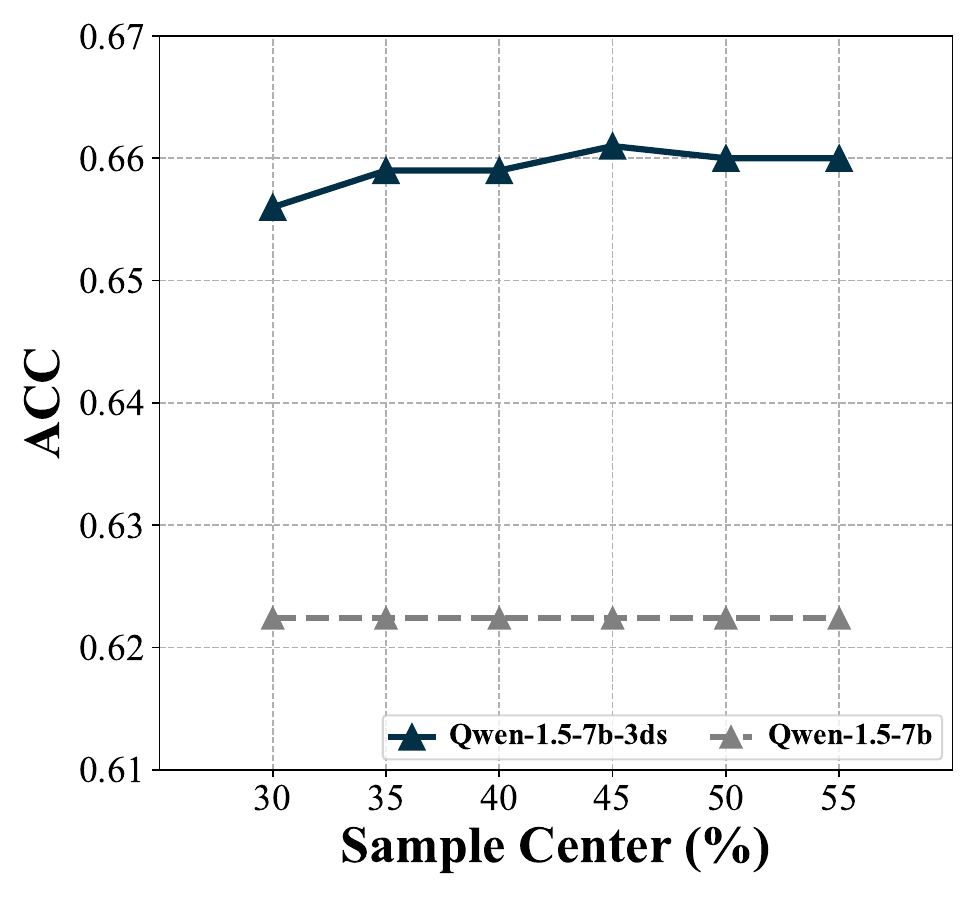}} 
    \caption{Impact of difficulty thresholds on model performance. Varying difficulty thresholds affect the accuracy of the models across different center percentile of selected difficulty range (\%).}
    \label{fig:difficulty_thresholds_plots}
\end{figure}

We conduct sliding-window experiments, varying the selection difficulty ranges ($\sigma\pm25\%$), to investigate how training data difficulty affects the model's medical domain fine-tuning. As shown in Figure~\ref{fig:difficulty_thresholds_plots}, the model's performance improves as difficulties increase, reaching a peak before declining. This pattern further highlights the importance of selecting data that best suits the model's learning capacity. Training on overly simple data limits improvements, while training on excessively difficult data impedes effective learning.

\begin{table}[ht]
\centering
\resizebox{\linewidth}{!}{
\begin{tabular}{lcccc}
\toprule
\textbf{Model} & \textbf{Metric Pair} & \textbf{Pearson $r$} & \textbf{Spearman $\rho$} \\
\midrule
\multirow{3}{*}{Qwen2.5-7B} 
& Instr.~vs Conf.~ & 0.0272 & 0.1458 \\
& Instr.~vs Corr.~ & -0.0446 & -0.0946 \\
& Conf.~vs Corr.~ & 0.0359 & 0.0270 \\
\midrule
\multirow{3}{*}{LLaMA3-8B} 
& Instr.~vs Conf.~ & -0.0367 & 0.1188 \\
& Instr.~vs Corr.~ & 0.0400 & 0.2736 \\
& Conf.~vs Corr.~ & 0.0589 & 0.0959 \\
\bottomrule
\end{tabular}
}
\caption{Correlation analysis of difficulty metrics.}
\label{tab:metric_correlation}
\end{table}

\begin{table}[ht]
\centering
\resizebox{\linewidth}{!}{
\begin{tabular}{lccc}
\toprule
\textbf{Model} & \textbf{Per Metric} & \textbf{3-Way Intersection} & \textbf{Ratio} \\
\midrule
Qwen2.5-7B & 34,715 & 9,744 & 28.1\% \\
LLaMA3-8B & 19,963 & 5,701 & 28.6\% \\
\bottomrule
\end{tabular}
}
\caption{Sample overlap across difficulty metrics.}
\label{tab:metric_overlap}
\end{table}

\subsection{Independence of Difficulty Metrics}

The three difficulty metrics, \textit{Instruction Understanding}, \textit{Response Confidence}, and \textit{Response Correctness}, are all derived from model perplexity. To validate these metrics provide sufficiently independent signals and contribute complementary information to selection, we examine the independence of the metrics. We computed Pearson and Spearman correlations between each pair of metrics on the medical dataset using Qwen2.5-7B and LLaMA3-8B. As shown in Table~\ref{tab:metric_correlation}, correlations are consistently low, indicating that the metrics capture complementary aspects of difficulty. We further examined the overlap of samples within the intermediate range of each metric. As shown in Table~\ref{tab:metric_overlap}, the small intersection of samples confirms that the metrics are not redundant and the combined selection meaningfully filters the data space. Overall, both correlation and overlap analyses validate the complementarity of our decomposed metrics, explaining why their joint use leads to stronger domain adaptation performance.

\section{Generalization to Other Domains}

\begin{table}[ht]
\centering
\begin{tabular}{lcc}
\toprule
\textbf{Method} & \textbf{Accuracy} & \textbf{Std. Dev.} \\
\midrule
No SFT & 57.77 & 0.25 \\
\midrule
 +Random & 73.40 & 0.80 \\
 +IFD & 60.30 & 2.62 \\
 +\M & \textbf{76.13} & 0.80 \\
\bottomrule
\end{tabular}
\caption{Accuracy (\%) comparison on law domain.}
\label{tab:legal_results}
\end{table}

\begin{table}[ht]
\centering
\resizebox{\linewidth}{!}{
\begin{tabular}{lccccc}
\toprule
\textbf{Method} & \textbf{Overall} & \textbf{Bio} & \textbf{Physics}  & \textbf{Philosophy} \\
\midrule
No SFT & 61.94 & 77.08 & 56.60 & 67.52 \\
\midrule
+Random & 63.25 & 75.69 & 57.45 & 69.45 \\
+IFD & 62.91 & 79.17 & 57.02 & 66.88 \\
+\M & \textbf{65.08} & \textbf{81.25} & \textbf{59.57} & \textbf{74.79} \\
\bottomrule
\end{tabular}
}
\caption{Accuracy (\%) comparison on MMLU.}
\label{tab:mmlu_results}
\end{table}

While our pilot study and main experiments focus on adapting LLMs to the medical domain using Chinese-language data, we note that our \M~is intrinsically domain-agnostic. To validate its cross-domain generalization ability, we conduct additional experiments on LLaMA3-8B-Instruct on the law and general domains.

\subsection{Law Domain}
On the law domain, experiments are done on an English-language dataset CaseHOLD~\cite{CaseHOLD10.1145/3462757.3466088}, with details of experiment setups introduced in Appendix~\ref{appendix:legal_domain_experiment}. We compare \M~with random selection and a strong baseline IFD. The results in Table~\ref{tab:legal_results} demonstrate that \M~consistently outperforms baselines in terms of accuracy, achieving an average accuracy of 76.13\% with low variance. These results suggest that our model-centric data selection \M~is effective for specialized domains beyond healthcare.

\subsection{General Domain}
To further assess cross-domain generalization, we benchmark \M~on the English MMLU~\cite{mmlu-hendrycksmeasuring} dataset, which contains tasks from a wide range of domains including natural sciences, humanities, and social sciences. Details of experiment setups are introduced in Appendix~\ref{appendix:legal_domain_experiment}. Table~\ref{tab:mmlu_results} reports both overall accuracy and representative subject results. \M~achieves the highest overall accuracy (65.08\%), outperforming both Random and IFD. Moreover, it yields consistent gains across heterogeneous subjects such as College Biology, Conceptual Physics, and Philosophy. These improvements highlight that \M~not only generalizes to another specialized domain but also scales to a broad, high-resource, multi-domain benchmark, reinforcing its robustness and wide applicability.

\section{Related Work}
\label{gen_inst}

\paragraph{Data Selection for LLM Training}
Data selection for LLM training has been widely explored. Some works~\citep{das2023deft} focus on diversity via statistical clustering or core-set selection, but often overlook data quality and risk introducing noise that hinders training. 
To address quality concerns, some works employ external evaluators like proprietary LLMs~\citep{chen2023alpagasus,liu2023deita,wettig2024qurating} or reward models~\citep{du2023mods,liao2025LearNAT} to select high-quality data. However, due to distributional gaps between external evaluators and the target model, data labeled as high-quality may still contain redundant or conflicting information, limiting its effectiveness. 
Another line of work leverages internal signals from the model itself, such as perplexity~\citep{marion2023less}, gradients~\citep{xia2024less}, or derived metrics like data learnability~\citep{zhou2023lobass}, instruction following difficulty~\citep{li2024quantity,li-etal-2024-superfiltering} and V-usable information~\cite{v-u-icl-lu2023measuring,v-usable-ethayarajh2022understanding}. While these signals provide more direct insights into the model's understanding of data, they typically offer only coarse difficulty estimates, failing to capture different aspects of data complexity or account for the model's generation behavior. Their focus on the hardest data also risks overwhelming the model. Though related to active learning~\cite{yoo2019learning_loss_for_active_learning,karamcheti2021mind_your_outliers,mindermann2022prioritized} in challenges and insights, LLM data selection differs in scale and objectives. In this work, we focus exclusively on data selection tailored for LLMs. 
We note that existing data selection methods for LLMs mainly focus on pre-training, general instruction-tuning (transforming a base model into a chat model), or task-specific fine-tuning. 
In contrast, data selection for domain adaptation fine-tuning remains underexplored, where unique challenges lie in selecting data that best elicit the model's diverse domain abilities. To bridge this gap and overcome the limitations of current methods, we introduce a novel model-centric data selection framework and provide fine-grained analysis of data difficulty, enabling better aligned data selection for LLM domain adaptation fine-tuning.

\paragraph{Data Learnability in LLM SFT}
LLMs encounter significant challenges when learning unfamiliar or complex knowledge during supervised fine-tuning, particularly when the data was not encountered during pre-training, which can impede domain adaptation fine-tuning. \citet{gekhman2024does} found that models acquire new factual knowledge slowly during SFT, especially when the information diverges from their pre-existing understanding, leading to a higher risk of hallucinations. \citet{ren2024learning} further shows that when the knowledge introduced during Instruction Fine-tuning significantly differs from what was learned in pre-training, the model struggles to integrate it, causing performance degradation. This highlights the difficulty models face in using pre-training knowledge to understand new concepts. \citet{kang2024unfamiliar} also emphasizes that unfamiliar examples during fine-tuning increase the likelihood of hallucinations, suggesting that high-difficulty data can destabilize the model and negatively impact its ability to adapt to new domains. Together, these findings underscore the risks associated with fine-tuning on excessively difficult data, which can undermine model performance in domain-specific tasks.

\section{Conclusion}
In this paper, we highlight the importance of selecting data aligned with the model's distribution for LLM domain adaptation fine-tuning through a pilot study. To this end, we propose a two-stage model-centric data selection framework \M. Stage\#1 explicitly aligns data with the LLM’s knowledge through prompt-driven selection. The Stage\#2 implicitly aligns data via difficulty decomposition. Leveraging Instruction Understanding, Response Confidence, and Response Correctness difficulties calibrated by attention-based importance weighting, \M~effectively models the LLM's implicit distribution and selects data well-matched to its learning capacity. 
Extensive experiments on multiple medical and legal tasks show significant performance gains, demonstrating \M's effectiveness and generalization ability.
Overall, we offer a path toward more efficient LLM domain adaptation fine-tuning. Future work will explore extending the framework to more domains and refining training strategies based on difficulty metrics for broader applications.
\section*{Limitations}
Due to time and resource constraints, we have only validated our method in the medical, law and general domains. The results show that \M~is domain-agnostic and adaptable to other fields. However, further experiments may still be needed to fully verify its generalization. \M~requires the model to rate the entire training set and perform inference on the selected subset. Although in experiments, we utilize VLLM to accelerate the process, it still involves certain computational costs. \M~performs data selection prior to fine-tuning. Considering that the model’s evaluation of data difficulty may evolve during training, future research should explore dynamic selection that adapts to the model's changing state. Additionally, data filtered out is currently discarded. Future work should consider integrating mechanisms such as human-in-the-loop validation or strategies to recover potentially relevant and valuable data from the discarded pool. Finally, considerations for social bias and fairness issues are discussed in Appendix~\ref{appendix:bias_and_fairness}.

\section*{Acknowledgments}
This work is supported by the Capital’s Funds for the Health Improvement and Research (No. 2024-2G-4106) of China.


\bibliography{0_ACL_main}

\newpage
\appendix
\section{Pseudo Codes of \M}
\label{appendix:pseudo_codes}

We provide the pseudo codes of \M~in Algorithm~\ref{algo:method}.

\begin{algorithm}[h]
\SetAlgoLined
\KwIn{Full dataset $\mathcal{X}$, model $M$, scoring threshold $\theta$, difficulty calculation functions $\text{D1}, \text{D2}, \text{D3}$, percentage thresholds $p_1, p_2, p_3$, sampling budget $k$}
\KwOut{Selected data subset $\mathcal{S}$}

\textbf{Stage\#1:  Prompt-Driven Data Selection} \\
Initialize $\mathcal{X}_1 \gets \emptyset$ \\
\ForEach{$x \in \mathcal{X}$}{
  Get score $s_x \gets M(\text{prompt}, x)$ \\
  \If{$s_x \geq \theta$}{
    Add $x$ to $\mathcal{X}_1$ \\
  }
}

\textbf{Stage\#2: Decomposed Difficulty-based Data Selection} \\
Initialize $\mathcal{S} \gets \emptyset$ \\
Compute $\text{D1}(x), \text{D2}(x), \text{D3}(x)$ for all $x \in \mathcal{X}_1$ \\
Set $\tau_1, \tau_2, \tau_3$ based on percentiles $p_1, p_2, p_3$ of $\text{D1}, \text{D2}, \text{D3}$\\
\ForEach{$x \in \mathcal{X}_1$}{
  \If{$\tau_1^{\text{low}} \leq \text{D1}(x) \leq \tau_1^{\text{high}}$ \textbf{and} 
      $\tau_2^{\text{low}} \leq \text{D2}(x) \leq \tau_2^{\text{high}}$ \textbf{and} 
      $\tau_3^{\text{low}} \leq \text{D3}(x) \leq \tau_3^{\text{high}}$}{
    Add $x$ to intermediate set $\mathcal{S}_\text{mid}$ \\
  }
}

Apply K-Center sampling on $\mathcal{S}_\text{mid}$ to select $k$ diverse data points \\
Return final selected subset $\mathcal{S}$

\caption{Model-Centric Data Selection Framework}
\label{algo:method}
\end{algorithm}

\section{Datasheet for Medical Domain Adaptation Fine-Tuning Dataset}
\label{appendix:train_dataset}

\textbf{Data statistics}

The statistics of the training dataset and the test dataset are shown below. The use of the test datasets complies with their respective licenses.

\begin{table}[h]
\centering
\begin{tabular}{lc}
\hline
\textbf{Dataset}              & \textbf{Size (K)} \\ \hline
medtalk\_singleround           & 177               \\
medknowledge\_KG              & 796               \\
medknowledge\_webqa            & 360               \\
medtask\_promptcblue           & 82                \\
qa\_website                   & 490               \\ \hline
\textbf{Total}                & \textbf{1905}      \\ \hline
\end{tabular}
\caption{Training Dataset Statistics}
\label{tab:train_dataset_statistics}
\hfill

\centering
\begin{tabular}{llc}
\hline
\textbf{Dataset} & \textbf{Type}        & \textbf{Size}     \\ \hline
CMB-Exam         & multiple-choice      & 11200                 \\ 
MMCU-Med     & multiple-choice      & 2819                      \\ 
CMB-Clin         & open Q\&A              & 208                \\ \hline
\end{tabular}
\caption{Test Dataset Statistics}
\label{tab:test_dataset_statistics}
\end{table}

\textbf{What is the primary purpose of creating this dataset?}

This dataset was created to construct a large-scale medical domain instruction-following fine-tuning dataset. The primary purpose is to support the adaptation of large language models (LLMs) to the medical domain by providing diverse and comprehensive training instances. By integrating heterogeneous data sources, including doctor-patient dialogues, medical knowledge bases, and various medical tasks formulated into the instruction-output format, the dataset aims to enhance the ability of LLMs to perform effectively across a wide range of real-world medical scenarios. It is designed to address the unique challenges of the medical domain, such as specialized terminology, complex reasoning, and context-sensitive responses, thereby enabling LLMs to better meet the demands of healthcare applications.

\textbf{What are the specific components of the dataset, and how were they constructed or sourced?}

Our dataset integrates multiple open-sourced medical instruction fine-tuning datasets from diverse sources, along with doctor-patient dialogue data extracted from medical consultation websites and a variety of medical tasks reformulated into the instruction-output format, as detailed in Table \ref{tab:train_dataset_statistics}. \textbf{Medtalk\_singleround} originates from open-sourced doctor-patient question-and-answer datasets, including CMedQA2~\citep{cmedqa2} and Health-Care-Magic\footnote{\url{https://www.kaggle.com/datasets/gunman02/health-care-magic}}.  
\textbf{Medknowledge\_KG} is built from the Online Medical Knowledge-Based Data in Huatuo26M~\citep{li2023huatuo26mlargescalechinesemedical}, which is derived from the extensive medical literature data provided by the Chinese Medical Association.  
\textbf{Medknowledge\_webqa} includes knowledge-driven, open-ended question-and-answer pairs in the medical domain, sourced from~\citep{medllmdata2023}.  
\textbf{Medtask\_promptcblue} combines the promptCBLUE dataset~\citep{zhu2023promptcblue} with additional data converted into the instruction-output format from the CBLUE benchmark~\citep{zhang-etal-2022-cblue}.  
\textbf{QA\_website} contains authentic doctor-patient dialogue data collected from the online platform of a collaborating hospital. Examples from these datasets are shown in Table \ref{tab:qa_table}.

\textbf{Are the data sources legal? How are privacy and ethical considerations addressed?}

The dataset is derived from carefully selected sources, including publicly available datasets and data crawled from the website of a collaborating hospital. Explicit permission was obtained from the collaborating hospital for the use of the crawled data, and all data have been anonymized to ensure that no personal information is exposed. Additionally, the hospital’s website provides open-access data, complying with relevant legal and ethical standards. This ensures the legality and security of the data while addressing privacy and ethical concerns.

\textbf{What are the potential risks and limitations of this dataset?}

The dataset has certain inherent risks and limitations that should be acknowledged. First, as the data is collected from diverse sources, it may contain noise or inconsistencies, which could affect the quality and reliability of downstream applications. Additionally, since the dataset is derived from Chinese text corpora, including medical advice and Q\&A exchanges, its content may be culturally and regionally specific, making it more suitable for East Asian populations. As a result, the medical recommendations and insights in the dataset may not generalize well to other demographic or cultural contexts.

To address these issues, users should carefully evaluate the dataset's suitability for their intended applications and, if necessary, consider adapting the data to align with broader use cases. Moreover, noise reduction and validation techniques can be employed to improve data quality and reliability in specific tasks.

\textbf{What is the usage case for this dataset?}

This dataset is primarily intended for instruction fine-tuning of large language models (LLMs), as already utilized in this study. Practitioners can use it to fine-tune LLMs to adapt to the medical domain, as well as to enhance its medical abilities in general fine-tuning. Additionally, the dataset may be useful for more specific tasks, such as fine-tuning for sub-tasks in the dataset.

\textbf{What is the distribution method and maintenance plan for this dataset?}

The dataset is distributed as an open-source resource at \url{https://drive.google.com/drive/folders/1SfrwQkDrQJ8i_EIqfc2Di0Xa5Y5pzY9H}, allowing researchers and developers to access and utilize it freely under the specified license. We are committed to the ongoing maintenance of the dataset. If any errors or inaccuracies are identified, particularly those related to medical knowledge, we will promptly update the dataset to correct such issues, removing erroneous data as necessary. Additionally, we will continue to provide updated documentation to ensure the dataset's effective use. While the dataset is stable at present, users are encouraged to provide feedback or suggest improvements, and we will consider updates based on user input or evolving needs in the field. This ensures that the dataset remains reliable and beneficial for the community.

\begin{table*}[h]
\centering
\begin{tabular}{lp{5.5cm}p{6.5cm}}
\hline
& \textbf{Medtalk\_singleround} & \textbf{English translation} \\ \hline
\textbf{Question} & \begin{CJK}{UTF8}{gbsn} 医生请问怀孕时母亲得了甲亢会遗传个孩子么？之前得过甲亢怀孕时又复发了但是没有吃药想知道宝宝会不会被遗传？\end{CJK} & Q: Doctor, can hyperthyroidism during pregnancy be inherited by the baby? Mine recurred during pregnancy, but I didn’t take medication. Will my baby be affected? \\
\textbf{Answer}  & \begin{CJK}{UTF8}{gbsn} 甲亢与遗传有一定的关系，但不是都会遗传的，不要担心。建议平时合理安排饮食，需要高热量高蛋白质高维生素和低碘的饮食。精神要放松。适当休息，避免重体力活。\end{CJK} & A: Hyperthyroidism has some genetic links but is not always inherited. Don’t worry. Maintain a balanced diet rich in calories, protein, and vitamins, but low in iodine. Stay relaxed, rest adequately, and avoid strenuous activities. \\ \hline
& \textbf{Medknowledge\_KG} & \textbf{English translation}  \\ \hline
\textbf{Question} & \begin{CJK}{UTF8}{gbsn} 颜面部凹陷的手术治疗有些什么？\end{CJK} & Q: What surgical treatments are available for facial depression? \\
\textbf{Answer}   & \begin{CJK}{UTF8}{gbsn} 自体颗粒脂肪移植；自体脂肪移植；自体脂肪干细胞移植；自体脂肪颗粒移植。\end{CJK} & A: Autologous micro-fat grafting, autologous fat transplantation, autologous stem cell-enriched fat grafting, and autologous fat particle transplantation.\\ \hline
& \textbf{Medknowledge\_webqa} & \textbf{English translation}  \\ \hline
\textbf{Question} & \begin{CJK}{UTF8}{gbsn} 请描述寄生虫病的治疗方法。 \end{CJK} & Q: Please describe the treatment methods for parasitic diseases. \\
\textbf{Answer}   & \begin{CJK}{UTF8}{gbsn} 以消灭寄生虫为主，根据虫种采用最有效的驱虫药物。在感染较重而寄主较衰弱时，可给予支持疗法，有外科并发症时应及时进行外科处理。\end{CJK} & A: The primary goal is to eliminate the parasites by using the most effective antiparasitic drugs based on the specific type of parasite. In cases of severe infection where the host is weakened, supportive therapy may be provided. Surgical intervention should be performed promptly if complications arise. \\ \hline
& \textbf{Medtask\_promptcblue} & \textbf{English translation} \\ \hline
\textbf{Question} & \begin{CJK}{UTF8}{gbsn} 问诊对话的实体抽取：两个礼拜片感冒了，现在还有点咳嗽，会不会和这个有关？选项：具体的药物名称，医疗操作，医学检查检验，症状。\end{CJK} & Q: Entity extraction in diagnostic dialogues: "Caught a cold two weeks ago, still have a bit of a cough now—could it be related?" Options: specific medication names, medical procedures, medical tests and examinations, symptoms.\\
\textbf{Answer}   & \begin{CJK}{UTF8}{gbsn} 上述句子中的实体包含：症状实体：感冒，咳嗽。 \end{CJK} & A: The entities in the above sentence include: Symptom entities: cold, cough. \\ \hline
& \textbf{QA\_website} & \textbf{English translation} \\ \hline
\textbf{Question} & \begin{CJK}{UTF8}{gbsn} 每天下午低烧三十六七℃，有时胸闷，没有咳嗦，盗汗，乏力的，有没有得肺结核的可能？\end{CJK} & Q: Low-grade fever of 36-37°C every afternoon, occasional chest tightness, no cough, night sweats, or fatigue—could this indicate a possibility of tuberculosis?\\
\textbf{Answer}   & \begin{CJK}{UTF8}{gbsn} 你这个体温其实从临床上来讲，不算是低烧，一般来讲，37度二以上才算是低热，所以说你这个跟集合的关系不是特别大的，你倒是可以看一下有没有病毒感染的可能，再一个，有没有新冠的问题？\end{CJK} & A: From a clinical perspective, this temperature doesn't qualify as a low-grade fever—typically, temperatures above 37.2°C are considered low-grade. Therefore, its connection to tuberculosis is unlikely. However, you might want to check for the possibility of a viral infection or consider whether it could be related to COVID-19.\\ \hline
\end{tabular}
\caption{Examples For various type dataset}
\label{tab:qa_table}
\end{table*}

\section{K-Center Sampling Algorithm}
\label{appendix:kcenter}
In our data selection framework, K-Center sampling is employed to ensure diversity within the selected instruction fine-tuning data. After filtering based on difficulty levels, we obtain an intermediate set $\mathcal{S}_{\text{mid}}$, composed of data points within a moderate difficulty range. The K-Center sampling is then applied on $\mathcal{S}_{\text{mid}}$. Specifically, the process works as follows:

\textbf{1. Embedding Generation:} For each data sample, the instruction part is encoded into an embedding using the LLM. We extract the last hidden states of the LLM and compute the average across all tokens in the sequence to form a fixed-size embedding vector. These embeddings represent the semantic content of the instruction.

\textbf{2. K-Center Sampling:} Using these embeddings, the K-Center sampling algorithm selects \( k \) data points in a greedy manner. The goal is to maximize the minimum distance between any pair of selected data points, ensuring that the sampled data points are as distinct as possible. This promotes diversity in the selected dataset and minimizes the risk of overfitting to similar data points.

The pseudo codes of this greedy K-Center sampling process are shown in Algorithm~\ref{algo:k-center}:

\begin{algorithm}[htbp]
\SetAlgoLined
\KwIn{Intermediate set $S_{mid} = \{s_1, s_2, \dots, s_n\}$, model $M$, data budget $k$}
\KwOut{Final selected set $\mathcal{S}$}

\textbf{Step 1: Encode data in $S_{mid}$ using model $M$}\;
\ForEach{$s_i \in S_{mid}$}{
    Encode $s$ using $M$ to obtain the embedding $e_s$ \;
}

\textbf{Step 2: Run K-Center greedy algorithm}\;
Initialize $\mathcal{S} \gets \emptyset$ \;
Initialize $\text{min\_distances} \gets \infty$ \;
\For{$i = 1$ \KwTo $k$}{
    \eIf{$\mathcal{S} = \emptyset$}{
        Select $s_j \in S_{mid}$ randomly and add it to $\mathcal{S}$ \;
    }{
        $min\_distances_j = \min_{s_i \in \mathcal{S}} \|e_{s_j} - e_{s_i}\|_2, \quad \forall s_j \in S_{mid} \setminus \mathcal{S}$\;
        
        Select $s^* = \arg\max_{s_j \in S_{mid} \setminus \mathcal{S}} min\_distances_j$\;
        
        Add $s^*$ to $\mathcal{S}$\;
    }
}
\Return{$\mathcal{S}$}
\caption{Greedy K-Center Sampling}
\label{algo:k-center}
\end{algorithm}

\section{Baseline Implementations}
\label{appendix:baseline}
Due to differences in task settings and datasets, we re-implement baselines using their publicly available codes. We adapt the selection strategies to our medical domain adaptation fine-tuning task. The re-implementation details are as follows. Our use of the code repositories complies with their respective licenses:
\paragraph{(1) Alpagasus:}\cite{chen2023alpagasus} We adopt the open-sourced implementation\footnote{ https://github.com/gpt4life/alpagasus}, officially verified by the original authors. Given the scale of the full training set, applying GPT-4 annotation to the entire set would incur substantial financial cost due to API usage. Constrained by our budget, we randomly sample 70K training samples and assess their quality using the provided prompt with GPT-4o. From data scoring above the default threshold of 4.5, we randomly select 5K samples.
\paragraph{(2) DEITA:}\cite{liu2023deita} We utilize the official implementation from the public GitHub repository\footnote{https://github.com/hkust-nlp/deita} and directly download their trained data quality and complexity scorers from HuggingFace\footnote{https://huggingface.co/hkust-nlp/deita-quality-scorer}\footnote{https://huggingface.co/hkust-nlp/deita-complexity-scorer} without modification. The scorers are applied to randomly sampled 70K training data. We then select the top 5K samples with the highest scores in both quality and complexity. 
\paragraph{(3) IFD:}\cite{li-etal-2024-superfiltering,li2024quantity} The Instruction Following Difficulty (IFD) method begins by calculating the instruction-following difficulty scores for each data point through model forward propagation. Given that our full domain dataset consists of over 1.9 million samples, performing this step on the entire dataset would be computationally prohibitive. Therefore, we randomly sample 60K samples from the training set, an amount comparable to the dataset size used in our \M~after Stage\#1. We compute IFD scores for this subset, and, following the recommendations in the original paper, select the samples with highest scores. The data budget is constrained to 5K samples, ensuring consistent with our main experimental setup.
\paragraph{(4) MoDS:}\cite{du2023mods} For the MoDS baseline, We follow the original paper's implementations, using the reward model \texttt{reward-model-deberta-v3-large-v2}\footnote{https://huggingface.co/OpenAssistant/reward-model-deberta-v3-large-v2} to score the full dataset. We then obtain samples with scores above 0.5, yielding a subset of 120k high-quality data samples. From this subset, we apply K-Center sampling to select 2k seed samples for model warm-up training. Subsequently, the trained model perform inference on the 120k high-quality subset, and these predictions are rescored using the same reward model. Data samples where model's generated answers score below 0 are deemed necessary and are combined with the seed samples. From this merged set, we randomly select 5K samples as the final training data, and train models from scratch on this final data.
\paragraph{(5) LESS:}\cite{xia2024less} The LESS method involves constructing a gradient library based on the original data, which incurs significant computational costs, particularly for the large dataset like ours. Similarly, we sample 60k data points to compute the gradients. Unlike the original LESS method that targets specific downstream tasks and uses samples from the targeting dataset to construct a validation set, our domain adaptation fine-tuning scenario does not involve fixed downstream tasks. Therefore, we randomly selected an additional 100 samples from the training set as the validation set. Then we run the provided codes and select 5K training samples.

\section{Implementation Details}
\label{appendix:hyperparameters}

The difficulty thresholds in our experiments are determined based on model performance on a hold-out CMB-validation set composed of 280 samples provided in the CMB benchmark~\cite{wang2023cmb}. All experiments are conducted using the PyTorch 2.4.0 in Python 3.9, on 8 NVIDIA H100 GPUs and an Intel(R) Xeon(R) CPU, with both training and inference performed using half-precision FP16 for efficiency. We employ the LoRA fine-tuning method, targeting all linear modules within the model, with a learning rate of \(5 \times 10^{-5}\), a batch size of 64, and a single epoch of training. The learning rate is scheduled using a cosine decay scheduler with a warmup ratio of 0.1. The LoRA rank is set to 8, and the input sequence length is cut off at 1024 tokens. DeepSpeed Zero-3 is used to optimize distributed training. For instruction scoring, response generation, and training, we use templates corresponding to each model, implemented through the llamafactory project~\cite{zheng2024llamafactory}.

Due to the high computational cost of training and testing LLMs, most existing instruction data selection studies conduct experiments with a single run for efficiency~\cite{li2024quantity,du2023mods}. We adopt this approach as well. However, to assess the reliability of our results, we perform the random selection experiment three times. The results show consistent performance with low variance (MMCU: 0.07; CMB 0.01 for \texttt{Qwen1.5-7B-Instruct}) and narrow error bars ($\pm$0.26 and $\pm$0.08 for \texttt{Qwen1.5-7B-Instruct}), demonstrating that our findings are statistically stable and reliable.

\section{Evaluation Metrics}
\label{appendix:metrics}
To evaluate the performance of LLMs on multi-task medical choice questions, we instruct the models to provide only the correct answer and adopt the widely-used metric, \textbf{Exact Match (EM)}, as recommended by prior work~\cite{f1,f2}. An answer is deemed correct under the EM metric if its form exactly matches all the correct answers listed in the ground truth. The EM score is computed as follows:
\begin{equation*}
EM = \frac{\text{Number of Correctly Matched Answers}}{\text{Total Number of Answers}} \times 100\%.
\end{equation*}

For open-domain medical Q\&A tasks, we employ \textbf{ROUGE-R}~\cite{xu2023contextaware,HyKGE} and \textbf{Bilingual Evaluation Understudy (BLEU)} to assess the quality of the LLMs' responses. 

\textbf{BLEU-N} Specifically, \textbf{BLEU-1} is used to measure answer precision, and \textbf{BLEU-4} evaluates answer fluency by considering higher-order n-gram consistency.
\textbf{BLEU} evaluates the similarity of generated responses to the ground truth using the following formula:
\[
\text{BLEU-N} = BP \cdot \exp\left(\frac{1}{N} \sum_{n=1}^{N} \log p_n\right),
\]
where \( p_n \) is the precision of \( n \)-grams, \( BP \) is the Brevity Penalty, calculated as:
    \[
    BP = 
    \begin{cases} 
    1, & \text{if } c > r \\
    \exp\left(1 - \frac{r}{c}\right), & \text{if } c \leq r
    \end{cases}.
    \]
    Here \( c \) is the length of the generated response, and \( r \) is the length of the reference response.

\textbf{ROUGE-R} quantifies the recall of retrieved knowledge in the LLMs' responses, emphasizing their ability to comprehensively cover the information relevant to the query. For a generated response \( R \) and a reference \( G \), ROUGE-R is computed as:
\[
\text{ROUGE-R} = \frac{|R \cap G|}{|G|},
\]
where \( |R \cap G| \) denotes the number of overlapping n-grams between the generated response and the reference, and \( |G| \) is the total number of n-grams in the reference.

During implementation, We use the 'rouge' package to calculate ROUGE scores and the 'nltk' module to compute BLEU scores (from BLEU-1 to BLEU-4), utilizing the smoothing function for BLEU and the default settings for ROUGE.

\section{Supplementary Experiments}
\label{appendix:supplementary_experiments}

\subsection{Win Rates Evaluation}
\label{appendix:win_rates}

When evaluating model performance on the open Q\&A dataset CMB-Clin, in addition to traditional metrics such as BLEU1, BLEU4 and Rouge scores, we conduct a pair-wise comparison to more thoroughly compare the fine-tuned models' medical analysis ability. In this experiment, we randomly sample 100 answers from each model and employ GPT-4o, a highly capable LLM, as the judge to determine which model generates a better answer. Below, we present the prompt used to instruct GPT-4o to compare answers from two models in this qualitative evaluation. To ensure a fair comparison and eliminate any possible positional bias in GPT-o4, we randomly assign the answers from each model as "Student 1" or "Student 2" throughout the experiment.

\begin{figure*}[!b]
\centering
    \subfloat{\includegraphics[width = 0.5\linewidth]{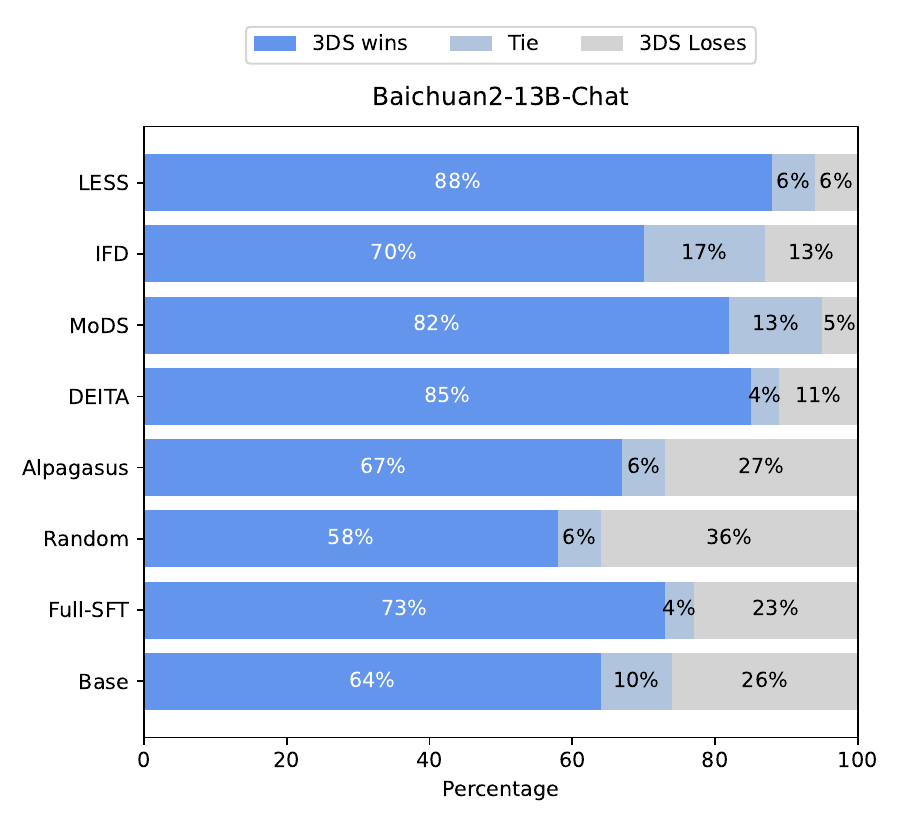}}
    \hfill
    \subfloat{\includegraphics[width = 0.5\linewidth]{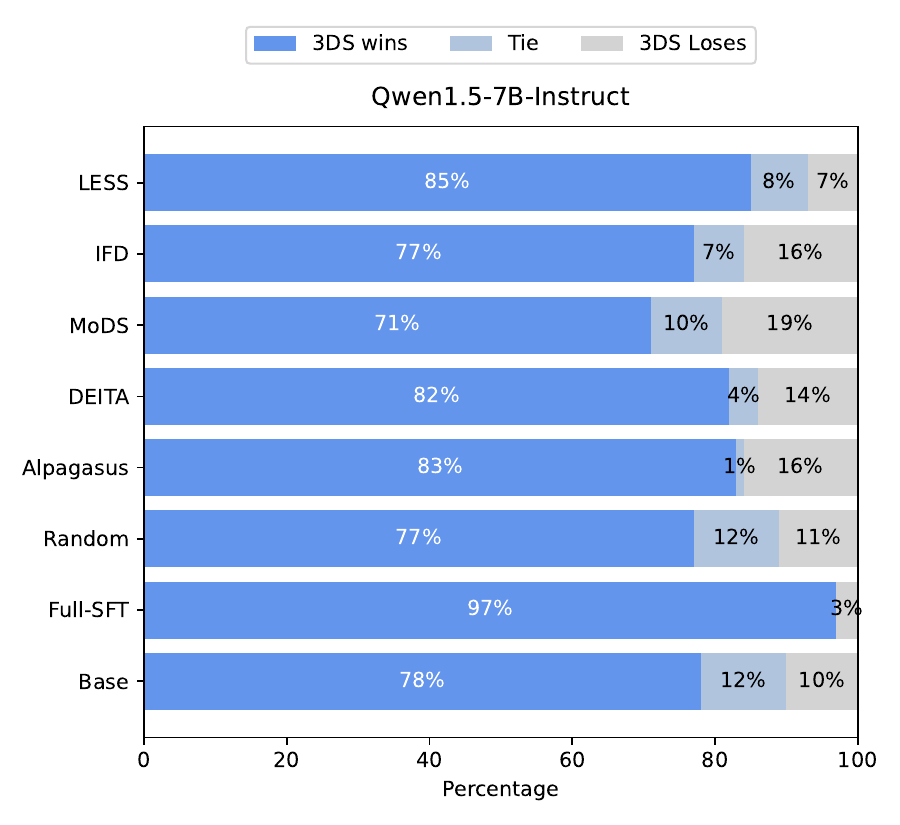}} 
    \hfill
    \subfloat{\includegraphics[width = 0.5\linewidth]{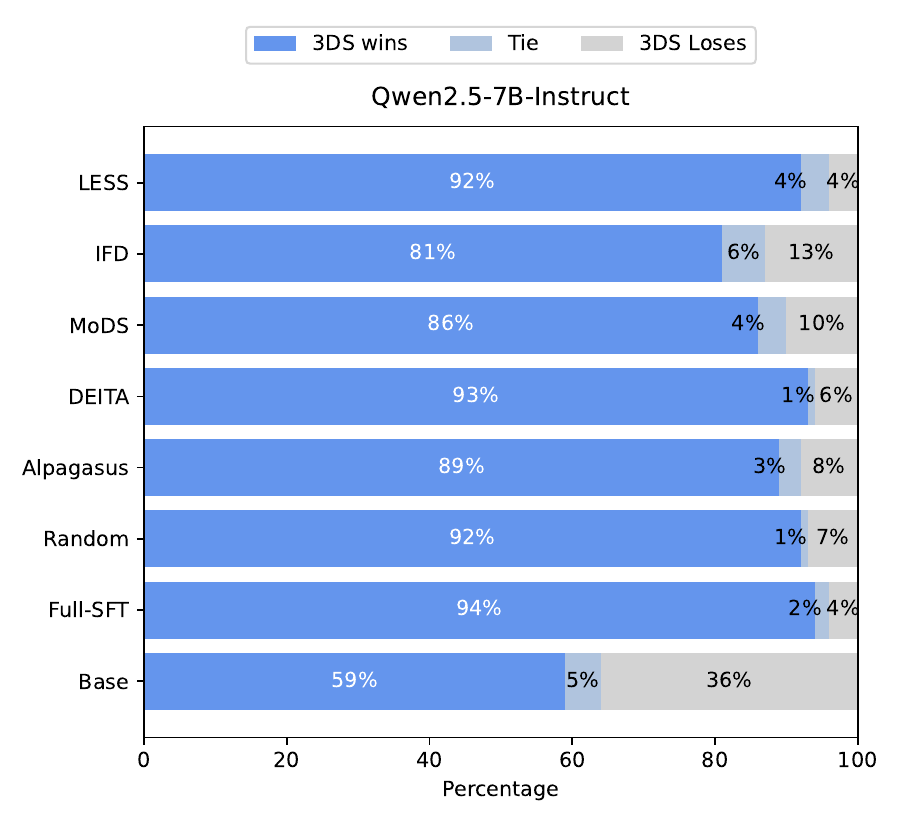}} 
    \hfill
    \subfloat{\includegraphics[width = 0.5\linewidth]{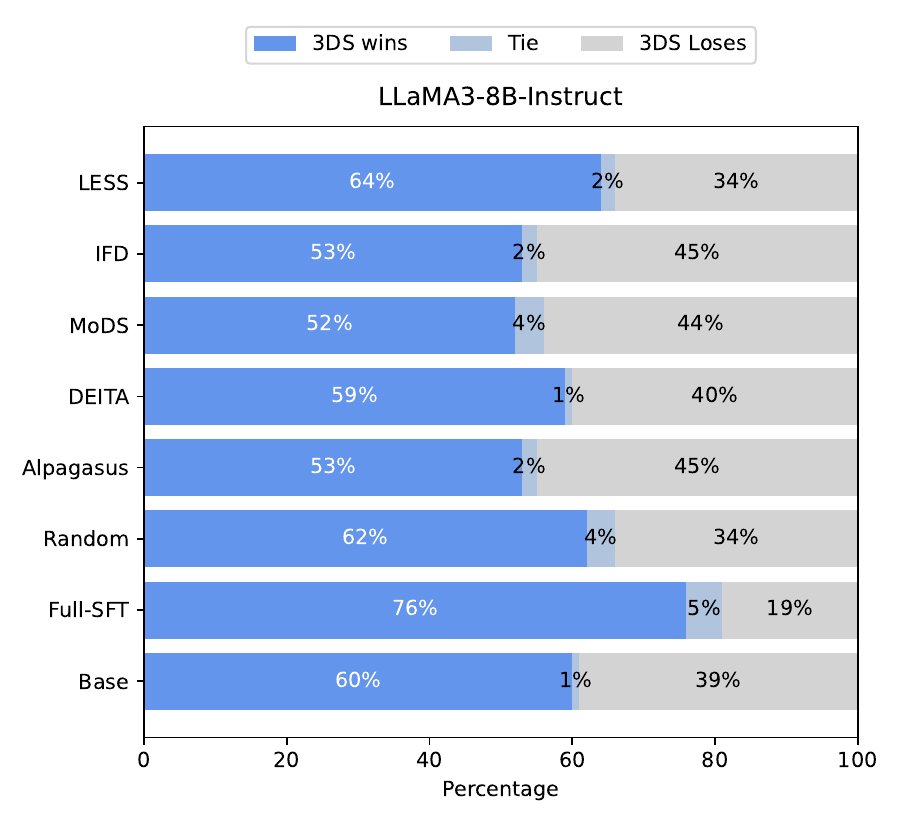}} 
\caption{GPT-4o judgment of CMB-Clin.}
\label{fig:gpt_judge}
\end{figure*}

Results shown in Figure~\ref{fig:gpt_judge} demonstrate that \M~exhibits substantially higher win rates compared to all other baselines. Notably, the larger and stronger models \texttt{Baichuan2-13B-Chat}, \texttt{Qwen1.5-7B-Instruct} and \texttt{Qwen2.5-7B-Instruct} generally show higher win rates compared to \texttt{LLaMA3-8B-Instruct}, which indicates that \M~also exhibits scalability. This evaluation provides qualitative evidence that \M~effectively enhances the model to deliver more clinically accurate outputs.


\begin{tcolorbox}[colback=lightgray!20,colframe=darkgray!80,title= CMB-Clin Evaluation Prompt,breakable]

You are now a medical expert guiding students in analyzing medical cases. You have two students, Student 1 and Student 2. You assess them through real medical case questions and choose the one with the best answer to become your assistant.
\newline
\newline
\textbf{\textit{[High-Quality Answer Criteria]}}\\
1. The answer should address the question directly and solve the problem posed.
\newline
\newline
2. The description of symptoms should be comprehensive and accurate, and the diagnosis should be the most reasonable inference based on all relevant factors and possibilities.
\newline
\newline
3. The treatment recommendation should be effective and reliable, considering the severity or stage of the condition.
\newline
\newline
4. The prescription should consider indications, contraindications, and dosages, being both effective and reliable.
\newline
\newline
\textbf{\textit{[Judgment Instructions]}}

Please compare the answers of Student 1 and Student 2. You need to tell me whether Student 1 is [better], [worse], or [equal] to Student 2. Compare their answers, refer to the question and the correct answer, and determine which one meets the given requirements more closely. Please only output one of the following: [Student 1 is better than Student 2], [Student 1 is worse than Student 2], or [Student 1 and Student 2 are equal]. Do not output any other words.
\newline
\newline
\textbf{\textit{[Case Example]}}

Here is the [Question]:\\
\texttt{<Insert medical question here>}\\

Here is the [Standard Answer]:\\
\texttt{<Insert standard answer here>}\\

Here is [Student 1]'s answer:\\
\texttt{<Insert Student 1's answer here>}\\

Here is [Student 2]'s answer:\\
\texttt{<Insert Student 2's answer here>}
\\
Please compare the two answers and give your judgment.
\end{tcolorbox}

\subsection{Comparison with Existing Medical LLMs}
\label{appendix:comparison_with_medical_LLMs}

\begin{table}[H]
\centering
\setlength{\tabcolsep}{1mm}
\begin{tabular}{c|cc}
\hline
\textbf{Model} & \textbf{CMB-Exam} & \textbf{MMCU-Med} \\ \hline
Baichuan2-13B-3DS & 47.37 & 51.08  \\
Qwen1.5-7B-3DS & 61.96 & 66.09 \\
Qwen2.5-7B-3DS & \textbf{79.06} & \textbf{85.70} \\
Meditron-7B & 11.20 & 12.16  \\
Huatuo-II-7B & 27.69 & 47.18 \\
Huatuo-II-34B & 59.54 & 66.10 \\
\hline
\end{tabular}
\caption{Performance comparisons with existing medical LLMs.}
\label{tab:comparison_with_medical_LLM}
\end{table}

To further validate the practical utility of \M, we conduct comparisons with existing medical LLMs. We compare \M~fine-tuned models to established medical LLMs, including open-source models MediTron~\cite{chen2023meditron70b} (7B version due to its similar size to Qwen models), and state-of-the-art Chinese medical LLMs HuatuoGPT-II-7B, and HuatuoGPT-II-34B~\cite{chen2024huatuogptii}, to see whether our framework can benefit the construction of medical LLMs. The results presented in Table~\ref{tab:comparison_with_medical_LLM} show that, MediTron-7B, as an English-based LLM, demonstrates limited performance on Chinese medical benchmarks. Huatuo-II-7B also falls short compared to similar-sized Qwen models. Huatuo-II-34B, with nearly five times the size of Qwen1.5-7B and Qwen2.5-7B, achieves only comparable performance. 

It is worth noting that the performance of fine-tuned models is closely tied to the capability of the base model, so relative improvements achieved through domain-specific fine-tuning are more important than absolute performance. Still, the strong performance of models fine-tuned with \M~validates its practical utility and efficiency for developing medical domain LLMs, paving the way for more building more powerful and advanced models in the future.

\subsection{More Results for Ablation on 3DS}
\label{appendix:more_results_for_3DS_ablation}

In the ablation studies in \ref{ablation_studies}, for CMB-Clin benchmark, we only report the ROUGE score. We provide BLEU-1, BLEU-4 scores and win-rates of the experiments in Table~\ref{tab:ablation_clin} and Table~\ref{tab:ablation_clin_win_rates}. Results are consistent with previous observations that the original \M~significantly outperforms ablation variants, supporting the validity of our designed two-stage framework and three data difficulty metrics.

\begin{table}[H]
\setlength{\tabcolsep}{1mm}
\centering
\resizebox{\linewidth}{!}
{
\begin{tabular}{c|cc|cc}
\hline
\multicolumn{1}{c|}{\textbf{LLM Turbo}} & \multicolumn{2}{c|}{\textbf{Baichuan2-13B-Chat}}& \multicolumn{2}{c}{\textbf{Qwen1.5-7B-Instruct}}\\
\hline
\textbf{Metric} & BLEU-1 & BLEU-4 & BLEU-1 & BLEU-4 \\
\hline
\textbf{w/o Stage\#1} & 14.13 & 29.60  & 15.50 & 31.94 \\
\textbf{w/o Stage\#2} & 20.56 & 46.86  & 21.55 & 47.39  \\
\textbf{Stage\#2 into Stage\#1} &  21.48 & 50.16 & \underline{21.73} & 52.27 \\
\hline
\textbf{w/o D1} & 22.55 & 51.75 & 24.14 & 55.12\\
\textbf{w/o D2} & 22.22 & 52.06 & 20.48 & 49.59\\
\textbf{w/o D3} & 20.86 & 49.40 & 22.27 & 50.18\\
\hline
\rowcolor[gray]{0.95}
\textbf{\M} &\textbf{24.15}  &\textbf{63.51} 
 &\textbf{24.40} &\textbf{60.32}  \\
\hline
\end{tabular}
}
\caption{Performance (BLEU-1, BLEU-4) on \textit{CMB-Clin} for ablation experiments. The best performance is highlighted in \textbf{bold}.}
\label{tab:ablation_clin}
\end{table}

\begin{table}[H]
\setlength{\tabcolsep}{1mm}
\centering
\resizebox{\columnwidth}{!}
{
\begin{tabular}{c|ccc|ccc}
\hline
\multicolumn{1}{c|}{\textbf{LLM Turbo}} & \multicolumn{3}{c|}{\textbf{Baichuan2-13B-Chat}}& \multicolumn{3}{c}{\textbf{Qwen1.5-7B-Instruct}}\\
\hline
\textbf{Metric}& Win & Tie  & Lose & Win & Tie  & Lose \\
\hline
\textbf{vs w/o Stage\#1} & 66.5 & 9.0 & 24.5 & 70.5 & 3.0 & 26.5\\
\textbf{vs w/o Stage\#2} & 66.0 & 15.5 & 28.5 & 66.0 & 5.5 & 28.5 \\
\textbf{vs Stage\#2 into Stage\#1}  & 63.5 & 18.0 & 18.5 & 54.5 & 2.5 & 43.0\\
\hline
\end{tabular}
}
\caption{Win-rates (\%) of GPT-4o judgment on \textit{CMB-Clin}, comparing \M~with stage ablation variants.}
\label{tab:ablation_clin_win_rates}
\end{table}

\begin{table}[H]
\setlength{\tabcolsep}{2mm}
\centering
\resizebox{\columnwidth}{!}
{
\begin{tabular}{c|cc|cc}
\hline
\textbf{Metric} & \multicolumn{2}{c|}{\textbf{CMB-Exam}} & \multicolumn{2}{c}{\textbf{MMCU-Med}} \\
\cline{1-5}
\textbf{LLM Turbo} & \textbf{K-Center (5K)} & \textbf{Full Pool} & \textbf{K-Center (5K)} & \textbf{Full Pool} \\
\hline
Qwen2.5-7B   & 79.06 & 79.14 & 85.70 & 86.24 \\
Qwen1.5-7B   & 61.96 & 62.17 & 66.09 & 66.02 \\
LLaMA3-8B    & 43.95 & 44.34 & 49.70 & 50.20 \\
Baichuan2-13B & 47.37 & 47.13 & 51.08 & 51.22 \\
\hline
\end{tabular}
}
\caption{Accuracy (\%) on CMB and MMCU-Med with and without K-Center sampling.}
\label{tab:kcenter_ablation}
\end{table}

\subsection{Ablation on Data Budgets}
\label{appendix:ablation on data budgets}
\begin{table*}[h]
\setlength{\tabcolsep}{1mm}
\centering
\resizebox{0.7\linewidth}{!}{
\begin{tabular}{cc|ccccc}
\hline
\textbf{Model}& \textbf{Dataset}  & \textbf{3K} & \textbf{4K}& \textbf{5K} &\textbf{6K} &\textbf{7K}  \\
\hline
\multirow{2}{*}{\textbf{Baichuan2-13B-Chat}}& CMB-Exam & 46.87 & 47.30 & \textbf{47.37}& 46.95 &46.98\\
&MMCU-Med & 48.67 & 49.91 & \textbf{51.08} & 50.16 & 50.27\\
\hline
\multirow{2}{*}{\textbf{Qwen1.5-7B-Instruct}}& CMB-Exam & 60.47 & 60.45 & \textbf{61.96} & 60.78 & 60.53\\
&MMCU-Med & 63.64 & 63.92 & \textbf{66.09} & 64.49 & 64.10\\
\hline
\end{tabular}
}
\caption{Performance comparison of models trained on different data budgets.}
\label{tab:ablation_data_budget}
\end{table*}

We conduct ablation experiments varying the selection data budgets. Results in Table~\ref{tab:ablation_data_budget} show that increasing the training data size initially boosts performance as the model learns to align with domain-specific knowledge. However, beyond a certain point (5K), performance degradations arise due to potential data redundancy and reduced diversity.

\subsection{Ablation on K-Center Sampling}
To further assess the role of K-Center sampling, we compare fine-tuning with and without this step. After filtering data to the moderate-difficulty range, samples may still be redundant, which can reduce efficiency. K-Center addresses this by selecting a diverse and representative subset, ensuring both data efficiency and stable sample sizes across datasets. We conduct an ablation study on two settings: (1) fine-tuning on 5K K-Center selected samples from the moderate-difficulty pool, and (2) fine-tuning on the entire pool without clustering. Results in Table~\ref{tab:kcenter_ablation} show that K-Center achieves comparable performance with significantly fewer samples, confirming its effectiveness in improving data efficiency.

\section{Threshold Selection Guidelines}
In \M's Stage\#2 Decomposed Difficulty-based Data Selection, data within a moderate difficulty range are selected. How to determine the optimal difficulty range is thus essential. We provide selection guidelines based on our experiments. We identify that the 25\%-75\% difficulty range is a robust choice. For model-specific optimization, we recommend this implementation procedure:
\begin{itemize}
    \item Model Capability Profiling: Conduct pre-fine-tuning validation to benchmark the model's baseline performance. Strong domain task performance suggests higher difficulty thresholds, while weaker models benefit from more conservative ranges.
    \item Hyperparameter Search: Implement search over potential ranges and select the values that yield the best performance on the validation set. This allows for adapting the difficulty range to the model's specific strengths and weaknesses.
\end{itemize}
While the exact thresholds are empirically tuned for performance optimization, the broader principle of selecting data of \textit{moderate difficulty} is supported by both theoretical intuition and recent findings~\cite{gekhman2024does,marion2023less,ren2024learning}.

\section{Cross Domain Experiment Details}
\label{appendix:legal_domain_experiment}
To assess the generalization ability of our model-centric data selection framework beyond the medical domain, we conduct experiments on law and general domains. 

On the law domain, \textbf{CaseHOLD} dataset~\citep{CaseHOLD10.1145/3462757.3466088} is utilized. This dataset consists of over 53,000 multiple-choice questions derived from U.S. court decisions. Each instance presents a case citation context along with five candidate legal holdings, of which only one is correct. The task simulates legal reasoning by requiring models to identify the option that best matches the cited precedent.

We follow a standard instruction-tuning setup by converting CaseHOLD into an Alpaca-style format. The \texttt{instruction} is fixed to a law domain-specific prompt:

\begin{tcolorbox}[colback=lightgray!20,colframe=darkgray!80,title= CaseHOLD Instruction,breakable]
As a law expert, please select the option that best matches the legal holding cited in the case. Answer with the option letter only (A/B/C/D/E).
\end{tcolorbox}

The \texttt{input} contains the case citation context and five formatted candidate holdings:

\begin{tcolorbox}[colback=lightgray!20,colframe=darkgray!80,title= CaseHOLD Input,breakable]
Case Citation Context: [citing\_context]
\newline
Options: A. [holding\_0] B. [holding\_1] \ldots E. [holding\_4]
\end{tcolorbox}

We fine-tune \texttt{LLaMA3-8B-Instruct} on 5K training samples selected from the CaseHOLD training set using three different strategies: \textbf{(1)} Random Selection, \textbf{(2)} IFD~\citep{li2024quantity}, a strong instruction filtering baseline, and \textbf{(3)} our proposed model-centric selection framework \textbf{\M}. All models are trained under the same hyperparameters, and each experiment is repeated three times with different random seeds. We report the mean accuracy and standard deviation on a selected 1K samples from the CaseHOLD test set.

On the general domain, MMLU~\cite{mmlu-hendrycksmeasuring} is utilized. MMLU spans $57$ subjects across diverse domains, including natural sciences, humanities, and social sciences. We fine-tune \texttt{LLaMA3-8B-Instruct} on subsets selected by different strategies (Random, IFD, and \M) from the 99k training set, and evaluate on the official MMLU test set.

\section{Data Evaluation Prompts}
\label{appendix:prompts}

\subsection{Data Quality Evaluation Prompt}
\label{appendix:quality_scoring_prompt}
In the pilot study and the first stage of \M, we utilize a prompt to instruct models to evaluate data quality on its internal knowledge. Inspired by existing works~\cite{chen2024huatuogptii,wang2023cmb,liu2023deita}, the model is asked to assess data quality across five dimensions: Instruction Complexity, Response Relevance, Response Thoroughness, Response Logic and Knowledge Richness. We provide the model with detailed scoring guidelines. The specific prompt used in this process is shown below.

\begin{tcolorbox}[colback=lightgray!20,colframe=darkgray!80,title= Quality Evaluation Prompt,breakable]
\label{tab:quality_prompt}
You are an AI assistant with medical expertise. Your task is to objectively assess the quality of the medical dialogue between the user and assistant based on your knowledge, and provide a score.
The data may consist of single or multi-turn dialogues. You should evaluate based on the complexity of the question, relevance of the response, thoroughness, logical coherence, and knowledge richness, and provide an overall score. Focus on medical-specific characteristics to ensure accuracy.
\newline
\newline
\textbf{\textit{[Evaluation Criteria]}}
\newline
\newline
1. \textit{Question Complexity}: Evaluate the complexity of the user's question. If the question requires deep understanding, reasoning, or medical knowledge, score above 80.
\newline
\newline
2. \textit{Response Relevance}: Assess if the assistant's response is directly aligned with the question. Score above 80 for responses tightly related to the question.
\newline
\newline
3. \textit{Response Thoroughness}: Check if the response thoroughly addresses the question with sufficient detail. A score above 80 reflects comprehensive answers.
\newline
\newline
4. \textit{Response Logic}: Ensure the response follows clear reasoning and logic. A score above 80 reflects well-structured reasoning.
\newline
\newline
5. \textit{Knowledge Richness}: Determine whether the response demonstrates rich, specialized medical knowledge. A score above 80 indicates depth and accuracy.
\newline
\newline

\textbf{\textit{[Scoring Guidelines]}}
\newline
\newline
[80-100]: Excellent. High complexity, thoroughness, relevance, logic, and knowledge richness, meeting medical standards.
\newline
\newline
[60-79]: Good. Strong performance but with minor deficiencies in logic or knowledge.
\newline
\newline
[40-59]: Fair. Noticeable issues such as unclear logic or insufficient depth.
\newline
\newline
[20-39]: Poor. Fails to properly address the medical issue or lacks substance.
\newline
\newline
[0-19]: Very Poor. Lacks relevance, logic, or medical knowledge.
\newline
\newline

\textbf{\textit{[Start Conversation]}}

Refer to the guidelines and score the following dialogue data based on the criteria. Follow the output format strictly: \\
\{score:\}

Dialogue: \\
\texttt{<qa\_pairs>} \\
Output:

\end{tcolorbox}

\subsection{Data Difficulty Evaluation Prompt}
\label{appendix:difficulty_scoring_prompt}
In the second empirical study, we prompt models to rate overall data difficulty according to its knowledge. The specific prompt used in this process is shown below.

\begin{tcolorbox}[colback=lightgray!20,colframe=darkgray!80,title= Overall Difficulty Evaluation Prompt,breakable]
\label{tab:difficulty_prompt}
You are a medical expert. I will provide you with an instruction related to the medical field. Based on your knowledge, please evaluate the difficulty of this instruction.
\newline
\newline
1. \textit{Medical Knowledge Complexity}: Does this instruction involve complex medical knowledge?
\newline
\newline
2. \textit{Reasoning Complexity}: Does answering this instruction require multi-step reasoning, integration of multiple sources of information, or handling clinical uncertainty?
\newline
\newline
3. \textit{Overall Challenge}: Considering the above factors, what is the overall difficulty of this instruction?
\newline
\newline
Based on these considerations, please provide a \textbf{comprehensive difficulty rating} from 1 (very easy) to 5 (very difficult). Only output the score; do not provide any explanation.
\newline
Instruction to evaluate:
\newline
\texttt{\{instruction\}}
\newline
\newline
Please return an integer between 1 and 5, representing the overall difficulty of the instruction for you. Only output the score and nothing else.
\end{tcolorbox}

\subsection{Decomposed Difficulty Prompts}
\label{appendix:decomposed_difficulty_prompts}
In the ablation study where we collapse Stage\#2 in \M~into Stage\#1, using prompts to instruct model to score the three decomposed data difficulties. The prompts utilized are listed below.

\begin{tcolorbox}[colback=lightgray!20,colframe=darkgray!80,title= Instruction Following Difficulty Prompt,breakable]

Based on your existing knowledge, evaluate the difficulty of understanding the following instruction.  
The higher the complexity and ambiguity of the instruction, the more difficult it is for the model to understand. Please provide a score between 0 and 1, where a higher score indicates that the instruction is more difficult for you to understand.
\newline
\newline
\textbf{Instruction to be evaluated:}  
\texttt{\{instruction\}}
\newline
\newline
Please return a real number between 0 and 1, representing the difficulty of understanding the instruction. Only output the score, and do not output anything else.
\end{tcolorbox}

\begin{tcolorbox}[colback=lightgray!20,colframe=darkgray!80,title= Response Confidence Difficulty Prompt,breakable]

Based on your existing knowledge, evaluate the difficulty of confidently and definitively providing the following evaluated response to the instruction.  
The more difficult it is to confidently provide this response, the higher the difficulty. Please provide a score between 0 and 1, where a higher score indicates greater difficulty in answering confidently.
\newline
\newline
\textbf{Instruction:}  
\texttt{\{instruction\}}
\newline
\textbf{Response to be evaluated:}  
\texttt{\{generated output\}}
\newline
\newline
Please return a real number between 0 and 1, representing the difficulty of confidently providing the response to the instruction. Only output the score, and do not output anything else.
\end{tcolorbox}

\begin{tcolorbox}[colback=lightgray!20,colframe=darkgray!80,title= Response Correctness Difficulty Prompt,breakable]

Based on the following instruction and the standard answer, evaluate the difficulty of providing the correct standard answer.  
If the instruction is complex or the answer requires high expertise, making it difficult to provide the correct answer, the difficulty will be higher. Please provide a score between 0 and 1, where a higher score indicates greater difficulty in providing the correct answer.
\newline
\newline
\textbf{Instruction:}  
\texttt{\{instruction\}}
\newline
\textbf{Standard Answer:}  
\texttt{\{output\}}
\newline
\newline
Please return a real number between 0 and 1, representing the difficulty of providing the correct answer. Only output the score, and do not output anything else.
\end{tcolorbox}

\section{Bias and Fairness Considerations}
\label{appendix:bias_and_fairness}
Fairness and bias are critical considerations, particularly in sensitive domains like healthcare. While our approach demonstrates promising results, its limitations and potential fairness implications must be acknowledged.
Our method employs the LLM to evaluate data quality and difficulty. Although the prompts and difficulty metrics are designed to be neutral, inherent model biases may still influence selection or be amplified through self-assessment, and the fairness impact of LoRA fine-tuning also requires further study~\cite{bui-von-der-wense-2024-tradeoff}. In downstream applications, biases introduced during SFT can be mitigated through prompt engineering, context augmentation and other techniques. Our design of self-assessment reflects a trade-off that prioritizes alignment and performance.
Potential bias also arises from training data composition, which predominantly consists of Chinese medical texts. While this dataset effectively reflects health conditions and medical practices of east Asian populations, it may limit generalizability to other populations.
Current LLM data selection methods generally prioritize difficulty, quality, or diversity, or diversity but rarely address fairness, safety, or truthfulness. Their effects on benchmarks such as SafetyBench \citep{zhang-etal-2024-safetybench} and TruthfulQA \citep{lin-etal-2022-truthfulqa}, remains underexplored. We therefore highlight fairness-aware and safety-aware data selection and fine-tuning as important directions for future work toward more equitable and reliable LLMs.

\end{document}